\documentclass[11pt]{article}

\usepackage[preprint]{acl}

\usepackage{times}
\usepackage{latexsym}
\usepackage{amsmath,amssymb}
\usepackage{enumitem}
\usepackage[T1]{fontenc}
\usepackage{xspace}

\usepackage[utf8]{inputenc}

\usepackage{microtype}
\usepackage{caption}
\usepackage{inconsolata}

\usepackage{graphicx}

\newcommand{\ec}{\emph{economicus}\xspace}

\title{Mind the (DH) Gap! A Contrast in Risky Choices Between\\Reasoning and Conversational LLMs}

\author{Luise Ge\thanks{Equal contribution}, Yongyan Zhang\footnotemark[1], Yevgeniy Vorobeychik \\
  Washington University in St. Louis \\
  \texttt{g.luise@wustl.edu, yongyan@wustl.edu, yvorobeychik@wustl.edu}}

\begin{document}
\maketitle
\begin{abstract}
The use of large language models either as decision support systems, or in agentic workflows, is rapidly transforming the digital ecosystem.
However, the understanding of LLM decision-making under uncertainty remains limited.
We study LLM risky choices along two dimensions: (1) prospect representation (based on an explicit representation or outcome history) and (2) decision rationale (explanation).
Our study, which involves 20 frontier and open LLMs, is complemented by a matched human subjects experiment, which 
provides one reference point, while an expected payoff maximizing rational agent model provides another.
We find that LLMs cluster into two categories: \emph{reasoning models (RMs)} and \emph{conversational models (CMs)}.
RMs tend towards rational behavior, are insensitive to the order of prospects, gain/loss framing, and explanations, and behave similarly whether prospects are explicit or presented via a history of outcomes.
CMs are significantly less rational, slightly more human-like, sensitive to prospect ordering, framing, and explanation, and exhibit a large description-history gap.
Paired comparisons of open LLMs suggest that a key factor differentiating RMs and CMs is training for mathematical reasoning.
\end{abstract}

\section{Introduction}

Large language models (LLMs) are increasingly used in decision support~\citep{benary2023leveraging,vrdoljak2025review} as well as agentic workflows that invoke tools and execute multi-step plans~\citep{moncada2025agentic,webb2025brain}. 
Consequently, LLMs are becoming significant \emph{economic decision-makers}, with uncertainty a central consideration.
Traditional computational paradigms for decision-making under uncertainty choose an option that maximizes expected utility or payoff~\citep{parkes2015economic}.
On the other hand, humans are known to systematically deviate from such behavior, and a host of mathematical decision models has been proposed to capture human risky choice~\citep{connolly2006regret,Kahneman1979PT,peterson2021using}.

However, LLMs are neither explicitly designed to maximize expected payoff, nor necessarily behave like humans.
Rather, their competence is an emergent property of scale and training paradigms including next-word prediction, instruction fine-tuning, human preference alignment, and mathematical reasoning~\citep{du2024understanding,guo2025deepseek,ouyang2022training}.
The net effect of this soup of training methods on LLM economic decision-making under uncertainty remains unclear.
In response, a literature has begun to emerge that aims to investigate LLM  decision-making, particularly under uncertainty~\cite{binz2023using,coda2024cogbench,jia2024decision,ross2024llm,payne2025analysis,wang2025prospect}.
However, there remain conflicting accounts, for example, of the extent LLMs are rational or align with prospect theory and human behavior~\citep{chen2023emergence,horton2023large,jia2024decision,ross2024llm,wang2025prospect,payne2025analysis}.
More fundamentally, given the broad variety of training paradigms, goals, and architectures, it is unclear whether a straightforward account of LLM economic decision making is even possible.

We study the~\textit{comparative} behavior of LLMs in a minimal, controlled two-option risky-choice benchmark, focusing on two underexplored dimensions: 1) how the prospects (uncertain options) are represented and 2) the impact of requesting LLMs to explain their decisions. The former is motivated by the description–experience (D–E) gap in human risky choice \citep{hertwig2004decisions,hertwig2009description,wulff2018meta}, which documents that people can choose differently when the same underlying prospects are learned through experience rather than read as explicit descriptions. The latter is motivated by evidence that verbalizing reasons can change human judgments and preferences \citep{festinger1962cognitive,shafir1993reason,wilson1991thinking}, and by the practical expectation that AI systems should be explainable \citep{ferrario2022explainability,zhao2024explainability}.

We consider 20 frontier and open LLMs, along with human subjects' choices over the same prospects and treatments.
LLMs are analyzed vis-a-vis two references: 1) the human subjects pool, and 2) the idealized %
\emph{economicus} rational agent. Within the scope of our experiments, our key findings are:

\begin{enumerate}[leftmargin=*,topsep=0pt,itemsep=-3pt]
    \item \textbf{LLMs cluster into two behavioral categories: \emph{reasoning models (RMs)} and \emph{conversational models (CMs)}.}  RMs are similar to the \ec\ while CMs are distinct from both the \ec\ and the human subjects pool.
    \item \textbf{All models exhibit a description-history (DH) gap}, an analogue of the decision-experience (DE) gap in which agents only see a history of past outcomes. This gap is considerable for CMs, but modest for RMs.
    \item \textbf{Explanations impact decisions for all models, though the impact is greater for CMs.} Surprisingly, they are typically more aligned with \ec ~under brief explanations than under no explanation or mathematical explanations.
    \item \textbf{Paired analysis of open models suggests that fine-tuning for mathematical reasoning is a key differentiator between RMs and CMs.} 
    Other parts of the training pipeline, however, appear to have limited impact.
\end{enumerate}

\smallskip
\noindent\textbf{Related Work.}
LLMs are increasingly studied as objects of behavioral analysis~\cite{binz2023using,coda2024cogbench,dillion2023can,hagendorff2023machine,hayes2024relative,horton2023large,ivanova2023running}. 
\citet{jia2024decision} analyze LLM decisions by eliciting choices and fitting prospect-theoretic models; subsequent work extends this by examining epistemic markers~\cite{wang2025prospect}, application contexts~\cite{payne2025analysis}, and persona matching~
\cite{liu2025evaluating}. 
\citet{horowitz2025llm} study a description-experience gap in LLMs through repeated choice, unlike the passive sampling used in our setting.
Apart from the bounded rationality focus, \citet{chen2023emergence} instead focuses on rationality (consistency with utility maximization) of GPTs directly (see also~\cite{jiang2025towards} for a survey) and \citet{mazeika2025utility} elicits the inherent value functions of various LLMs, while \citet{coda2024cogbench} provide a benchmarking suite for LLM decision making under risk.
We provide a broader discussion of related research in Appendix~\ref{sec:app_related_work}.
In general, existing work tends to focus narrowly on a single theory, relies on limited model coverage, overlooks the effects of choice representation and output format, or lacks matched human comparisons. Our work aims to address some of these limitations.

\section{Study Design}

We investigate how large language models (LLMs) make decisions when faced with uncertainty. To do this comprehensively, we examine 20 different LLMs (detailed in the Supplement, Appendix~\ref{subsec:llmslist}). Our selection covers two important goals: first, we include widely used frontier models; second, we use open-weight models of varying sizes and training stages to enable controlled comparisons.
We query all LLMs through a unified interface using a minimal instruction template (see Supplement, Appendix~\ref{subsec:prompts} for our prompts and an additional prompt sensitivity analysis). To measure behavior, we compare LLM responses against two references: human subject behavior and \ec, a risk-neutral expected payoff maximizer.

\subsection{Choices Among Prospects}
Our study uses three base prospect pairs as a structured testbed. These pairs span different outcome scales and are intentionally similar to (but not duplicates of) classic behavioral-economics stimuli \citep{kahneman2013prospect,peterson2021using}. This design reduces the risk of LLM memorization. See  Appendix~\ref{subsec:base_prospects} for details.

\noindent\textbf{Description-History Gap.}
Human decision-making differs depending on how information is presented. Research in the description–experience (DE) gap shows that people decide differently when given explicit information versus when they learn through experience. We adapt this insight to study LLMs. Rather than having LLMs interact dynamically with an environment (which would introduce confounds), we present prospects in two ways:
\begin{itemize}[topsep=0pt,itemsep=-5pt,leftmargin=*]
    \item \emph{Explicit Prospects.} Each prospect specifies exact payoff--probability pairs. For example: ``70\% chance of \$100, 30\% chance of \$0.''
    \item \emph{Implicit Prospects.} We present prospects as simulated histories -- sequences of payoffs that would result from repeated selection. For instance, a simulated history might show 15 instances of \$100 and 5 instances of \$0 from 20 trials, representing the same underlying distribution.
\end{itemize}

We call the behavioral difference between these presentations the \emph{description--history (DH) gap}. We test with simulated histories of 20 and 100 payoffs, with results aggregated across sample sizes (see Appendix~\ref{subsec:sample_size} for breakdowns by size).

\noindent\textbf{Decisions and Explanations.}
Our second key consideration involves the impact that requesting LLMs to provide a reason (explanation) has on its decisions.
We implement this using three prompt styles that request either 1) no explanation (output the choice only), 2) a one-sentence justification (\emph{short explanation}), or 3) a brief mathematical or reasoning-style justification (\emph{math explanation}).
See the Supplement (Appendix~\ref{subsec:prompts}) for details.

\subsection{Human Subjects Experiments}
\label{S:human}

To compare model behavior with human decision-making, we collected responses from 360 U.S.-based participants via Prolific.
Each participant selects among the same set of prospect pairs as LLMs
and is compensated at an average rate of \$24/hour.
In analysis, we treat humans as a single population distribution over choices.
This study was approved by the institutional IRB. We also conducted an initial attention check by excluding participants who spent an average of less than 8 seconds per question. As this exclusion did not lead to significant changes in the results, we report the analyses in the main text including all participants.

\subsection{Interpretable Models of Behavior}
\label{S:design}

To complement our analysis of the raw behavioral data, we make use of interpretable models of (LLM or human) behavior to obtain deeper insights into behavior and effective risk preferences.
We consider two parameterizations of prospect theory, both with four free parameters (presented in full in Appendix~\ref{subsubsec:pt_model}). The first is a standard formulation with parameters $\sigma$ (risk preference), $\lambda$ (loss aversion), $\gamma$ (probability weighting), and $\beta$ (decisiveness). However, because $\lambda$ and $\beta$ are not jointly identifiable in pure-loss prospects, we also consider an alternative specification as our primary model (a special case of the generalized power value model~\cite{peterson2021using}), replacing $\lambda$ and $\beta$ with  $\beta_{\text{gain}}$ and $\beta_{\text{loss}}$ for the gain and loss frames, respectively, while retaining $\sigma$ and $\gamma$.
In these models:
\begin{itemize}[leftmargin=*,topsep=0pt,itemsep=-2pt]
    \item Higher $\sigma$ implies risk seeking, $\sigma \rightarrow 0$ entails risk aversion, while $\sigma = 1$ is risk neutral. 
    \item Higher $\gamma$ implies underweighting of small probabilities, $\gamma \rightarrow 0$ indicates overweighting of these, and $\gamma = 1$ means no probability distortion. 
    \item Higher $\beta_{\text{gain}}$ (and $\beta_{\text{loss}}$) indicates greater determinism in the gain (and loss) domain, while $\beta \rightarrow 0$ implies essentially uniformly random choice. The ratio $\beta_{\text{loss}}/\beta_{\text{gain}}$ serves as a proxy for loss aversion: a ratio greater than 1 indicates higher sensitivity to utility differences in losses than in gains.
\end{itemize}
To avoid pathologies, we bound all parameters in $[0.01,1000]$.
An \ec\ then roughly corresponds to $\sigma = \gamma = 1$ and $\beta_{\text{gain}} = \beta_{\text{loss}} = 1000$.

\subsection{Quantifying Behavior}

\noindent\textbf{Querying and parsing.}
For each model and instantiated condition, we estimate a response distribution through repeated querying of 10 times under identical inputs.
We use temperature $T=1$ and a maximum generation length of 1024 tokens. We found that increasing the number of samples to 50 has minimal qualitative impact on the results, while being significantly more costly. 
If an LLM output does not contain a valid prospect choice,
we retain the raw output and mark the trial as invalid.
Choice rates are computed over valid trials.%

\noindent\textbf{Model Similarity and Goodness of Fit.}
Let $x$ refer to a \emph{pair of prospects, 
as well as a particular way of presenting these}, which includes: prospect order, framing (loss vs.~gain), nature of explanation requested, and prospect presentation (explicit vs.~implicit, and the length of the history for the latter).
$X$ denotes the set of all such contexts.

For a given context $x$ and a model $m$ (e.g., LLM, or a parametric model that we fit to data collected from LLM or human choices), we use $p_m(x) \in [0,1]$ to denote the fraction of times (probability) that $m$ selects the reference prospect (whichever we choose it to be) in this context.
For the \ec\, $p_e(x)$ is deterministic, while 
 $p_h(x)$ denotes the fraction of human subjects who selected the reference prospect.

For a pair of models $m$ and $m'$ (which can also refer to $h$ or $e$),
we can measure how well $m$ \emph{predicts} (or fits) the behavior of $m'$ in two ways.
The first is \emph{mean-squared error (MSE)}:
\[
\mathrm{MSE}(p_m,p_{m'})=\frac{1}{|X|}\sum_{x \in X} (p_m(x)-p_{m'}(x))^2.
\]
However, when MSE is non-zero, 
it may fail to capture an important property that decision patterns ``go together'', i.e., the extent to which an increase or decrease in $p_m$ \emph{across different contexts} is accompanied by a similar change in $p_{m'}$.
To capture this, we treat 
$p_m(x)$ as a random variable with context $x$ viewed as the associated outcome.
We can then measure similarity between a pair of models $m$ and $m'$ using \emph{Pearson correlation} between responses over all contexts $p_m(X)$ and $p_{m'}(X)$, $\mathit{Cor}(p_m(X),p_{m'}(X))$.
If $m$ is an interpretable parametric model fit to data, we evaluate both MSE and correlation on a held out (test) dataset that is distinct from the data on which the parameters of $m$ were fit.
Since the ability to capture the trends of behavior across contexts $x$ is essential to our analysis, we use correlation as the primary measure of both (a) similarity of pairs of models (e.g., LLMs), and (b) goodness of fit of a parametric model.
Nevertheless, we observe that both MSE and correlation typically lead to the same goodness-of-fit conclusions.

\noindent\textbf{Decisiveness and Consistency.}
In addition to model similarity and goodness of fit, two other behavioral traits we explore are \emph{decisiveness}, which captures the entropy in the decisions, and \emph{consistency}, of which we consider three forms.
The first is \emph{order consistency}, capturing the degree to which decisions tend to differ as a function of the order in which the prospects are presented.
The second is \emph{prompt consistency}, which measures the impact of explanation types on the nature of decisions.
Finally, we consider \emph{frame consistency}, which captures the impact of the gain vs.~loss framing of otherwise identical prospects on the decision.

Formally, decisiveness of a model $m$ (or a human $h$) is defined as $\frac{1}{|X|}\sum_{x \in X}\max\{p_m(x),1-p_m(x)\}$.
For consistency measures, let $S(x)$ define a collection of variations from the base context $x$ that we consider.
For example, in the case of ordering, $S(x)$ is comprised of before and after a given prospect is switched.
Order and prompt consistency is then defined as $1-\frac{1}{|X|}\sum_{x \in X} |\max_{x' \in S(x)} p_m(x') - \min_{x' \in S(x)} p_m(x')|$.  Frame consistency is defined as $1-\frac{1}{|X|}\sum_{x \in X} |p_m(gain) - (1- p_m(loss))|$.

\section{Behavior Patterns and Convergence}

\begin{figure}[h]
    \centering
    \includegraphics[width=\linewidth]{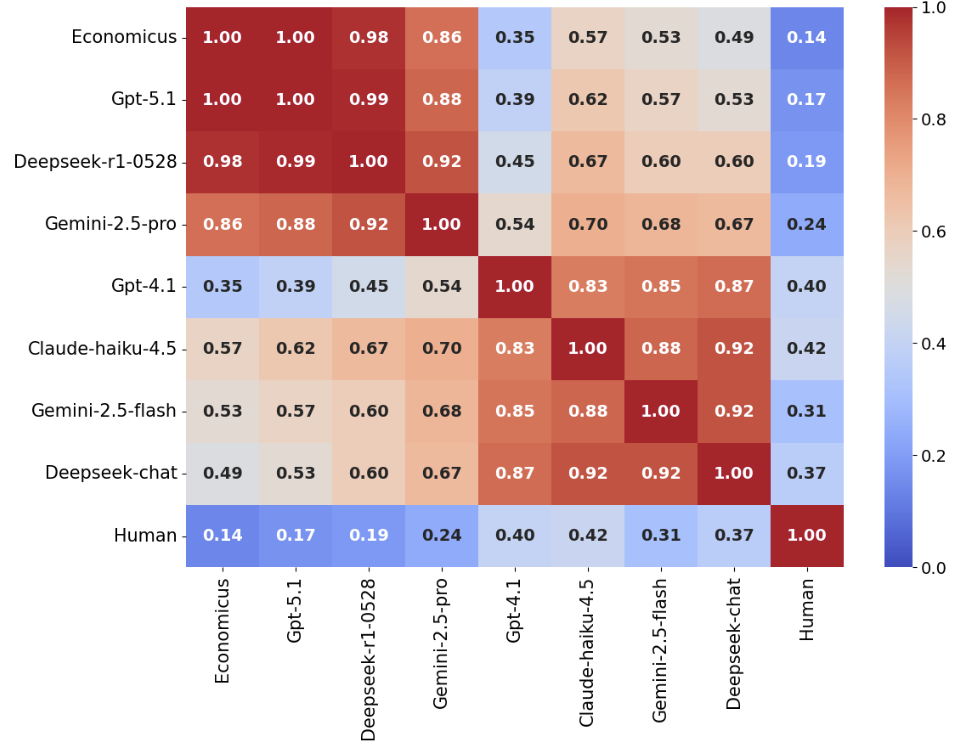}
    \caption{Correlation matrix involving (1) frontier LLMs (GPT-4.1 and 5.1, Gemini-2.5-Flash and Pro, DeepSeek-R1 and Chat, and Claude-Haiku-4.5), (2) \ec, and (3) human responses.}
    \label{fig:correlation-small}
\end{figure}

We begin our analysis by considering aggregate behavior for each LLM, as well as that of human and \ec\ references.
In Figure~\ref{fig:correlation-small} we present pairwise correlations between pairs of \emph{frontier} LLM models, as well as humans and \ec.

An immediate observation from the figure is \emph{the emergence of two LLM clusters}: one that includes the (frontier) \emph{reasoning models} GPT-5.1, DeepSeek-R1, and Gemini-2.5-Pro, and the other including their conversational counterparts (GPT-4.1, DeepSeek-Chat, and Gemini-2.5-Flash), as well as Claude-4.5-Haiku.
In particular, correlations among pairs of models \emph{within} each cluster are  $>0.8$, and in most cases $\approx 0.9$ or higher.
In contrast, correlations among pairs of models \emph{across} clusters tend to be considerably lower---typically $\approx 0.6$ or below.
This appears to be an example of  \emph{convergence} of frontier LLMs~\citep{smith2025comprehensive,zhou2025shared,mazeika2025utility}, albeit across two distinct lines that reflect their target use.
Henceforth, we thus distinguish between \emph{reasoning models (RMs)} and \emph{conversational models (CM)} in terms of their respective decision-making behavior under uncertainty.
This split into RMs and CMs is also observed in open models; see the full heatmap in the Supplement (Appendix~\ref{subsec:heatmaps}). %

Our next finding is that \emph{neither RMs nor CMs are particularly similar to human behavior}.
This observation is reflected in the remarkably low pairwise correlation in Figure~\ref{fig:correlation-small} between each frontier LLM and human: the highest is with the Claude Haiku model at only 0.42.
However, we do observe an intuitive pattern: RMs are considerably more different from human behavior than the CMs.

On the other hand, RMs are quite similar to the \emph{economicus} reference, with correlations ranging from 0.86 to 1.
In contrast, CMs exhibit low similarity to \emph{economicus}, with correlations below 0.5.
We also note that human behavior is essentially uncorrelated with \emph{economicus}, with correlation only 0.14.
This is likely a consequence of selecting prospects that most emphasize the deviation of human behavior from expected utility maximization.

To confirm that these structural differences are robust to sampling noise, we conducted an empirical bootstrap analysis: for each model and prospect pair, we resampled 1,000 synthetic datasets from a Binomial distribution parameterized by the observed choice frequency, and recomputed all behavioral metrics. The RM--CM clustering, as well as the deviation of human behavior from both, hold with $p < 0.001$ across all comparisons.

\begin{figure}[h]
    \centering

    \includegraphics[width=\linewidth]{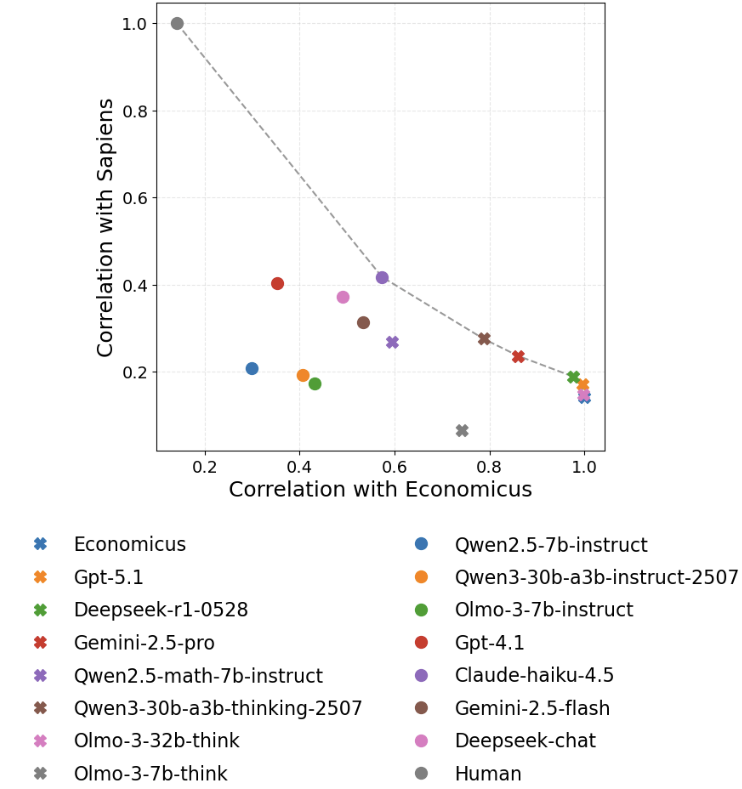}
    \caption{Pairwise correlation between each frontier model and both the human and \emph{economicus} references, shown as a 2D plot. Circles are CMs while x's are RMs.}
    \label{fig:map-2D}
\end{figure}

To simplify presentation, in Figure~\ref{fig:map-2D}, we introduce the \emph{HE 
representation}, plotting each model on a two-dimensional space anchored by human and 
economicus correlations (note humans have a non-zero correlation to the \ec). This visualization provides a compact, interpretable framework 
for understanding LLM behavior relative to two behavioral extremes.
The RM and CM clusters emerge here as well, with open models mapping predictably to each.
For example, Qwen2.5-7B-Instruct is a CM model, while its mathematical reasoning variant is clearly an RM model (being far more \ec-like and somewhat less human-like than the instruct version).
We see the same pattern with Qwen3-30B %
as well as with Olmo-3-7B (although here, the ``Think'' variant is slightly more human-like).
In general, we observe that RMs tend to be \emph{far more} \ec-like, and (usually) slightly less human-like compared to CMs.

The addition of open models also serves to provide a deeper insight into \emph{what features of the training paradigm lead to the CM-RM distinction}.
The key, it appears, is \emph{explicit mathematical reasoning}: models that are designed and trained for mathematical reasoning are consistently in the RM group, while those which are not fall into the CM group.

Moreover, the open models also allow us to observe the impact that model size has.
Specifically, we find that in general, larger models tend to be more human-like and more economicus-like compared to their smaller counterparts.
This can be viewed as a form of Pareto dominance, treating the Pareto frontier (the line in Figure~\ref{fig:map-2D}) as providing the best tradeoffs in the human-to-\ec\ similarity space.
Thus, smaller models are typically Pareto dominated by larger models. Moreover, an open, possibly smaller RM, can dominate a frontier CM (Qwen3-30B-thinking v.s. Gemini-2.5-Flash), suggesting that the size \emph{and} training techniques both play a significant role. 
We do find that all frontier RMs are at or near the Pareto frontier.

In our final aggregate analysis, we consider the \emph{decisiveness} and \emph{consistency} properties (i.e., relative invariance of decisions as we change the order and framing of prospects, as well as if we request an explanation) of LLMs, also comparing with the reference provided by the pool of human subjects.

\begin{figure}[h]
    \centering
    \includegraphics[width=0.9\linewidth]{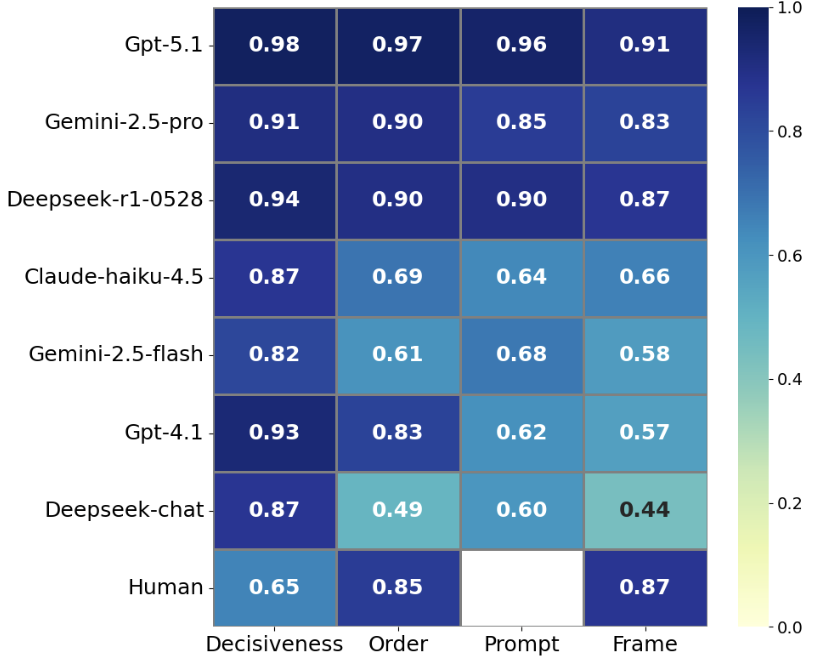}
    \caption{Consistency and decisiveness heatmap for frontier models and the human subjects.}
    \label{fig:c-d-small}
\end{figure}
The results provided in Figure~\ref{fig:c-d-small} for the frontier models again exhibit a clear distinction between RMs and CMs along each of these dimensions.
Actually, all LLMs appear to be rather decisive, especially as compared to the general human subjects pool, but RMs are nevertheless generally more decisive than CMs.
The difference in terms of consistency is even more substantial.
RMs are nearly order-invariant (just like \emph{economicus} would be), whereas CMs tend to be strongly influenced by the order in which the prospects appear.
CMs are also highly sensitive to the framing (gain vs.~loss) and prompt (nature of explanation requested; more on this in Section~\ref{S:explaining}), while RMs are not.
Remarkably, both in terms of order and frame sensitivity, \emph{humans are far more like RMs}.
For example, surprisingly, \emph{human behavior appears to be considerably less influenced by gain/loss framing than CMs}.

\section{Description-History Gap in LLMs}

In this section, we explore the \emph{description-history gap (DH gap)}---that is, the behavior discrepancies between explicit (description) and implicit prospects (experience history)---of LLMs and human subjects in the same contexts $x$. 

\begin{figure}[h]
    \centering
    \includegraphics[width=\linewidth]{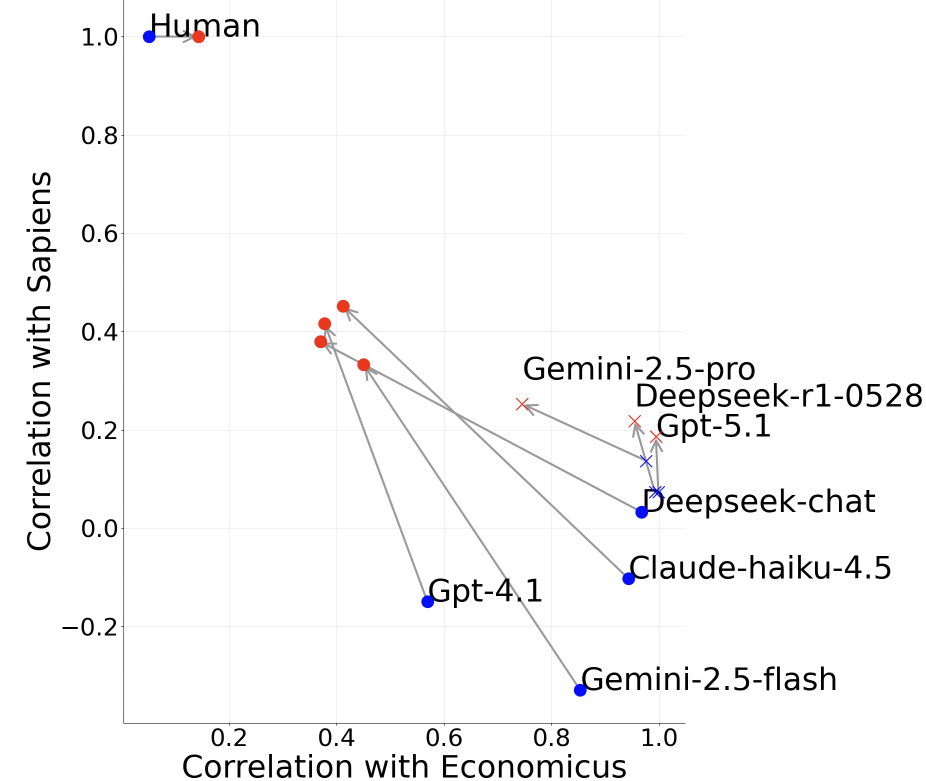}
    \caption{\emph{HE representation} of frontier LLMs (with human subjects as the reference). Change from explicit (blue) to implicit (red) prospects for each model is indicated by the arrows. RMs are x's and CMs are circles.}
    \label{fig:degap_closed}
\end{figure}
\begin{figure}[h]
    \centering
    \includegraphics[width=\linewidth]{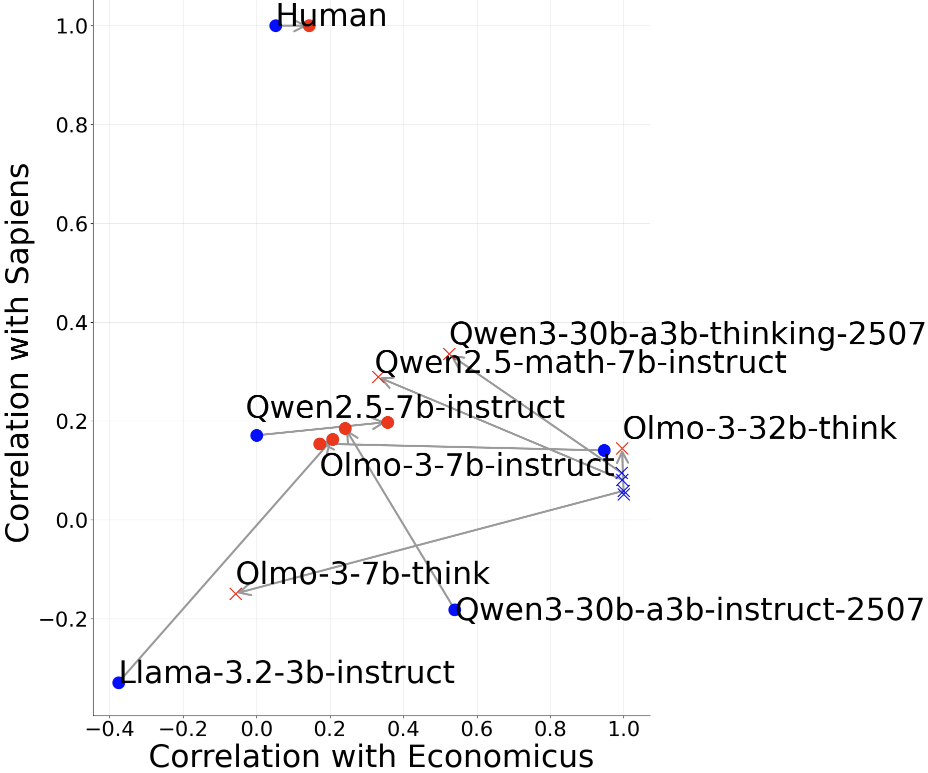}
    \caption{\emph{HE representation} of open LLMs (with human subjects as the reference). Change from explicit (blue) to implicit (red) prospects for each model is indicated by the arrows. RMs are x's and CMs are circles.}
    \label{fig:degap_open}
\end{figure}

The results, presented as the \emph{HE representation} of each model as well as the human subjects, are provided in Figure~\ref{fig:degap_closed} for the frontier LLM models and in Figure~\ref{fig:degap_open} for open LLMs.
Consider first the frontier LLMs, for which the observations are more crisp (Figure~\ref{fig:degap_closed}).
An immediate and rather striking observation is that for \emph{all} models---whether RMs (x's) or CMs (circles)---the change from explicit to implicit prospects \emph{increases the similarity to human behavior, and lowers similarity to economicus}.
An even more striking observation is the difference in \emph{how much} the change to implicit prospects impacts RMs and CMs.
In the case of the former, the impact tends to be moderate.
For example, neither GPT-5.1 nor DeepSeek-R1 become especially less \ec-like when deciding from experience, though the change is more notable with Gemini Pro.
For CMs, however, the change is dramatic: they become significantly more human-like, and significantly less \ec-like.
Indeed, while RMs remain similar to one another in either setting, CMs occupy a relatively broad HE representation range under explicit prospects, but cluster closely with implicit prospects.
For reference, we also show the change for the human subjects, who, somewhat surprisingly, become \emph{more economicus-like} when prospects are presented \emph{implicitly}; this likely accounts for the increased LLM-to-human correlation for RMs with implicit prospects.

We can see some of the same trends with open LLMs (Figure~\ref{fig:degap_open}), but with rather substantive differences.
In particular, reasoning models no longer exhibit a small DH gap that we observed with frontier models: moving from explicit to implicit prospect representation, nearly all gaps are rather dramatic, with the bulk of the shift away from being \ec-like (i.e., to the left), although with a slightly more human-like shift as well.
In addition, we can observe a non-trivial model size effect: the shifts are sharper, and differences from \ec-like behavior milder, with larger (30B and 32B) models than with smaller (7B) models.
Nevertheless, we generally still see the RM vs.~CM distinction, with the former models consistently to the right (more \ec-like) of the latter.

We note that all the models we tested have very long context windows. For example, DeepSeek-Chat has a 128K context limit. Even a 100-sample history represents only a small fraction of this available context, making it unlikely that the observed gap stems from context length limitations. This is further supported by results in the Appendix~\ref{sec:app_results} showing that many LLMs exhibit more rational behavior with longer (100-sample) histories than with shorter (20-sample) histories, providing strong evidence that context constraints are not the underlying cause of the DH gap.

\section{Explaining Decisions}
\label{S:explaining}

Besides asking LLMs to make autonomous decisions under uncertainty, it is quite natural to additionally request a rationale---an explanation---of their decisions.
However, at least when it comes to human decisions, providing such a rationale may itself impact the decision. Are LLMs similarly impacted when asked to provide an explanation?

\begin{figure}[h]
    \centering

    \includegraphics[width=\linewidth]{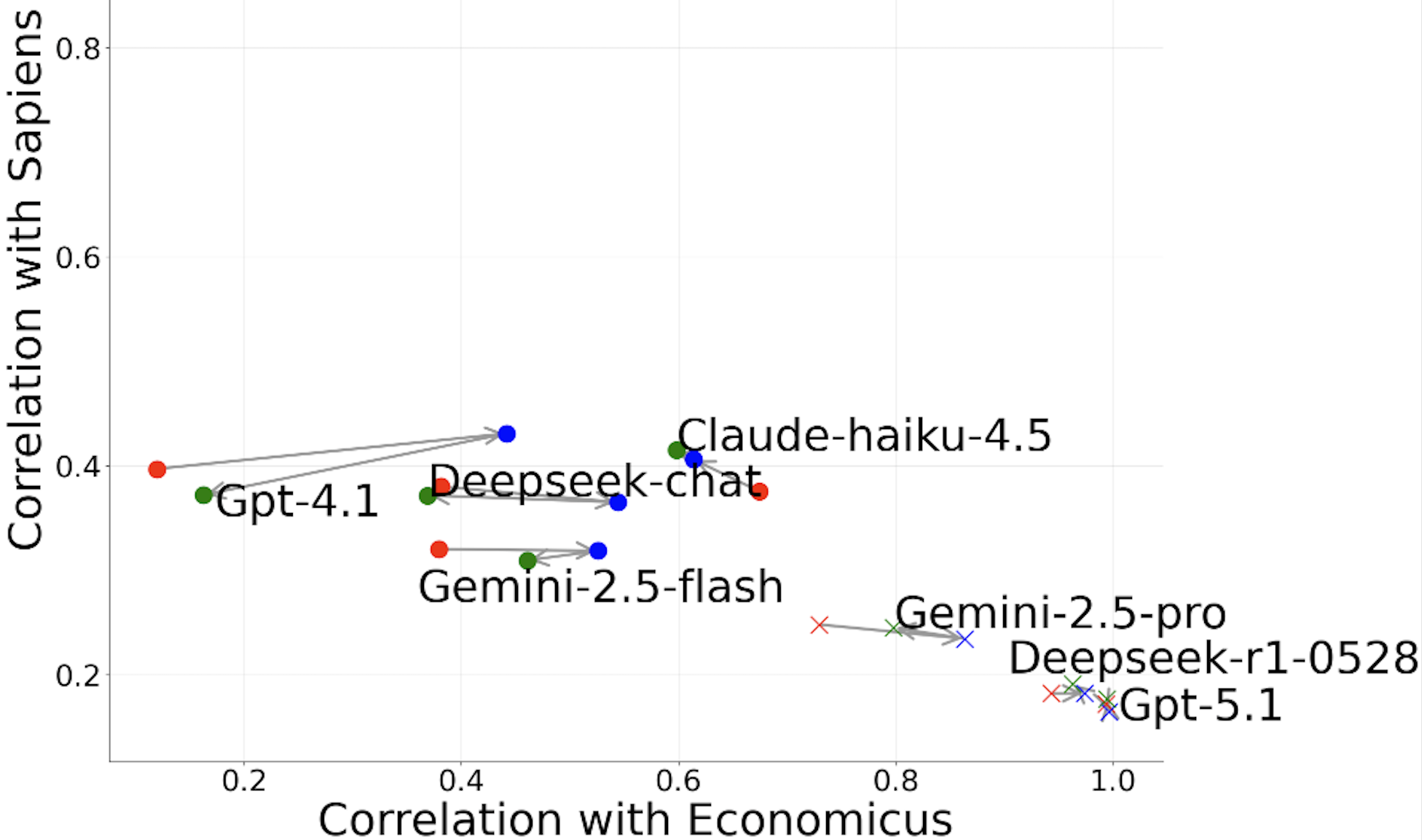}
    \includegraphics[width=\linewidth]{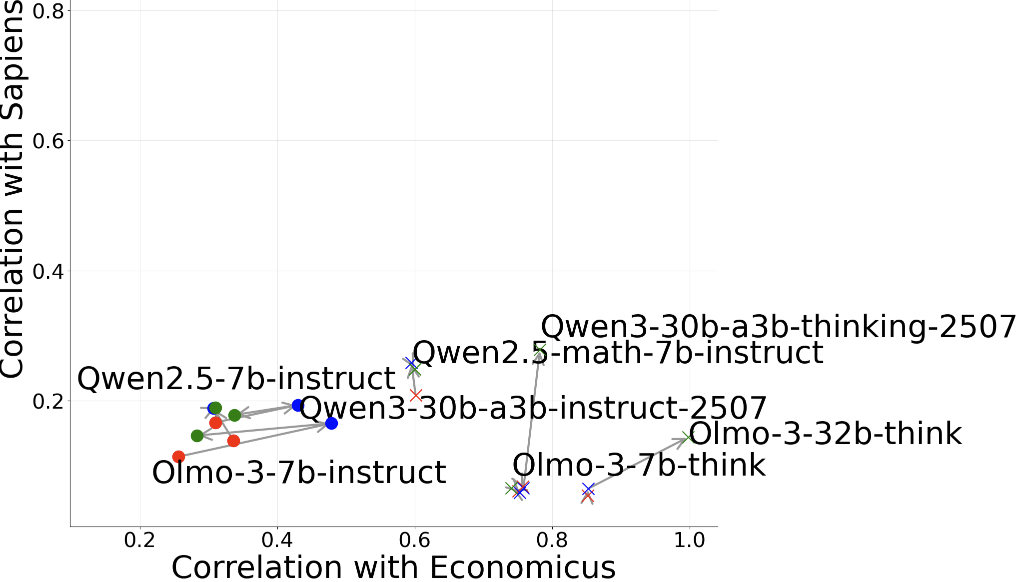}
    \caption{HE representation and the impact of explanations. Red: no explanation; Blue: one-sentence explanation; Green: math explanation.}
    \label{fig:explanation_closed}
\end{figure}
We investigate this by prompting LLMs for two kinds of explanations: a short (one-sentence) explanation and a full mathematical reasoning trace.
Figure~\ref{fig:explanation_closed} visualizes the impact of these two explanation modes. %
What emerges is a surprising pattern, captured by the sequence of arrows starting at ``no explanation'' (red), then pointing to ``short explanation'' (blue), and then to ``math explanation'' (green).
In particular, short explanations increase the degree of rationality (i.e., correlation with \ec) of most models, while asking for a longer rationale often \emph{reduces decision rationality in comparison with the former}.
Notably, this pattern is largely consistent for both frontier and open CMs, with Claude-4.5-Haiku somewhat of an outlier, whose \emph{both} types of explanations result in behavior that is less like \ec\ compared to no explanation provided. Trends in RM behavior across explanation modes are more heterogeneous: More than half of the models (GPT-5.1, DeepSeek-R1, Olmo3-7B-think, and Qwen2.5-Math-7B-Instruct) exhibit strong stability; Gemini-2.5-Pro follows a pattern similar to the CMs; In contrast, Qwen3-30B-Thinking and Olmo3-32B-Think show significant changes when prompted to provide a mathematical explanation, while no differences are observed between the no-explanation and brief-explanation modes. One possible explanation is that, because CMs are not trained for mathematical reasoning, prompting them to provide mathematical explanations may only induce verbosity but no improved rationality.

\section{Interpretable Models of LLM Behavior}

\subsection{Analysis of Frontier Models}

Our analysis thus far has considered the relationships among LLMs to one another, as well as to the behavior of human subjects and a rational risk-neutral utility maximizer (\ec).
However, such comparisons yield only relatively coarse insight into actual behavioral tendencies.
To dig deeper, we fit simple 4-parameter models of behavior to our data  (see Section~\ref{S:design}) for both LLMs and human subjects to offer interpretable
characterizations of the underlying  tendencies of LLMs that drive their decisions, contrasting these with both human subjects and \ec.
Here, we present the results for the generalized power value model; 
results for the alternative parameterization are provided in Appendix~\ref{subsec:old_pt_results}.

\begin{table}[h]
\centering
\begin{minipage}{0.5\textwidth}
\centering
\resizebox{\textwidth}{!}{%
\begin{tabular}{|l|c|c|c|c|c|c|}
\hline
\textbf{Model} 
& $\boldsymbol{\sigma}$ 
& $\boldsymbol{\gamma}$ 
& $\boldsymbol{\beta}_{\text{gain}}$ 
& $\boldsymbol{\beta}_{\text{loss}}$ 
& \textbf{Corr}
& \textbf{MSE}\\
\hline
DeepSeek-R1   & 0.86 & 0.81 & 832 & 711 & 0.98 & 0.007 \\
Gemini-Pro    & 0.81 & 0.84 & 815 & 857 & 0.96 & 0.014 \\
GPT-5.1       & 0.87 & 0.81 & 1000 & 802 & 1.00 & 0.001 \\
\hline
Claude Haiku  & 0.89 & 0.80 & 221 & 244 & 0.71 & 0.053 \\
DeepSeek-Chat & 0.86 & 0.82 & 181 & 149 & 0.77 & 0.040 \\
Gemini-Flash  & 0.93 & 0.89 & 27 & 42 & 0.87 & 0.015 \\
GPT-4.1       & 0.79 & 2.17 & 205 & 0.01 & 0.55 & 0.050 \\
\hline
Human         & 0.87 & 0.79 & 96 & 0.01 & 0.79 & 0.004 \\
\hline
\end{tabular}
}
\end{minipage}
\caption{Dual-Beta prospect theory parameters for decisions with explicit prospects.}
\label{T:explicit}
\end{table}

The results for explicit prospects are provided in Table~\ref{T:explicit} for the frontier models (corresponding confidence intervals are provided in  Appendix~\ref{subsec:ci_results}).
First, we observe that human $\sigma$ and $\gamma$ are comparable to those of most LLMs, suggesting similar risk preferences and probability weighting across agents. The key distinction lies in decisiveness: humans exhibit moderate $\beta_{\text{gain}}$ but near-zero $\beta_{\text{loss}}$, indicating they are far less consistent in the loss domain than in the gain domain.
We can also note that human behavior appears to be relatively predictable by the simple 4-parameter prospect theoretic model, with MSE $<0.01$ and correlation between the model and actual human behavior $>0.75$.

As we would expect, the frontier RMs exhibit behavior that is relatively close to rational (\ec): $\sigma$ and $\gamma$ are relatively close to 1, while both $\beta_{\text{gain}}$ and $\beta_{\text{loss}}$ are high.
RMs are also extremely predictable by simple parametric models, with nearly perfect correlation between predictions and behavior.
Moreover, even CMs' behavior in this setting appears to be close to \ec, with the substantive difference from RMs being considerably smaller values of $\beta_{\text{loss}}$ and $\beta_{\text{gain}}$ (lower level of determinism).
For the most part, CMs, too, are relatively well modeled by the simple 4-parameter models, with the exception of GPT-4.1, which has a lower correlation between predictions and behavior and a near-zero $\beta_{\text{loss}}$ --- resembling humans in being essentially random in the loss domain, though with high $\beta_{\text{gain}}$ in the gain domain.

\begin{table}[h]
\centering
\begin{minipage}{0.5\textwidth}
\centering
\resizebox{\textwidth}{!}{%
\begin{tabular}{|l|c|c|c|c|c|c|}
\hline
\textbf{Model} 
& $\boldsymbol{\sigma}$ 
& $\boldsymbol{\gamma}$ 
& $\boldsymbol{\beta}_{\text{gain}}$ 
& $\boldsymbol{\beta}_{\text{loss}}$ 
& \textbf{Corr}
& \textbf{MSE}\\
\hline
DeepSeek-R1   & 0.99 & 1.00 & 1000 & 346 & 0.98 & 0.009 \\
Gemini-Pro    & 0.98 & 0.99 & 147 & 100 & 0.94 & 0.025 \\
GPT-5.1       & 1.01 & 0.99 & 460 & 1000 & 0.98 & 0.010 \\
\hline
Claude Haiku  & 1.07 & 0.98 & 4.81 & 23 & 0.55 & 0.078 \\
DeepSeek-Chat & 0.83 & 0.72 & 11 & 0.01 & 0.45 & 0.048 \\
Gemini-Flash  & 1.57 & 0.82 & 2.61 & 14 & 0.58 & 0.047 \\
GPT-4.1       & 1.53 & 0.74 & 0.01 & 39 & 0.58 & 0.106 \\
\hline
Human         & 0.98 & 0.92 & 7 & 7 & 0.53 & 0.022 \\
\hline
\end{tabular}
}
\end{minipage}
\caption{Dual-Beta prospect theory parameters for decisions with implicit prospects.}
\label{T:implicit}
\end{table}

Turning next to implicit prospects (Table~\ref{T:implicit}), our first notable observation is that \emph{human behavior is now much closer to economicus}: $\sigma$ and $\gamma$ are now all relatively close to 1, and notably $\beta_{\text{gain}}$ and $\beta_{\text{loss}}$ are now roughly equal, indicating that the strong gain--loss asymmetry in decisiveness observed with explicit prospects disappears.
RMs largely remain close to \ec\ in this setting.
Moreover, all three RMs remain highly predictable with the simple 4-parameter models: correlation is near-perfect, while MSE is quite low.

In the case of CMs, implicit prospects induce considerable variation in behavior.
Claude Haiku and Gemini-Flash both show $\beta_{\text{loss}}$ several times larger than $\beta_{\text{gain}}$, indicating greater decisiveness in the loss domain, while GPT-4.1 shows an even more extreme version of this pattern ($\beta_{\text{loss}}=39$ vs.\ $\beta_{\text{gain}}=0.01$).
Most CMs exhibit some degree of probability overweighting ($\gamma < 1$), with the exception of Claude Haiku ($\gamma \approx 1$). All CMs have decisiveness that is now comparable to that of the human subjects pool, but several orders of magnitude below RMs.
Notably, frontier CMs in both explicit and implicit prospect settings are considerably less predictable with 4-parameter prospect theory models than RMs.

We note that these results differ in several ways from the alternative PT parameterization (Appendix~\ref{subsec:old_pt_results}). Most strikingly, the standard model fits extreme values of $\sigma$ and $\gamma$ for human subjects ($\sigma=0.04$, $\gamma=0.13$ for explicit prospects), corresponding to strong risk aversion and probability distortion that sharply distinguish humans from LLMs. 
Under the dual-beta specification, however, human $\sigma$ and $\gamma$ are comparable to those of most LLMs ($\sigma=0.87$, $\gamma=0.79$); the distinction shifts entirely to the decisiveness parameters. This discrepancy likely arises because $\lambda$ and $\beta$ are confounded in the loss domain (see Appendix~\ref{subsubsec:pt_model}), causing the standard model to compensate through distorted value function parameters. The standard model also exhibits instability in $\lambda$ and $\beta$ across RMs (e.g., $\lambda=1000$ paired with $\beta=1000$ for several models), further illustrating the identifiability issue. Despite these differences, both parameterizations achieve similar goodness of fit and yield qualitatively consistent conclusions regarding the RM--CM distinction.

\subsection{Analysis of Open Models: the Impact of Training and Scale}

Our final analysis makes use of open models to investigate the impact of training paradigms on the effective LLM behavior. Table~\ref{tb:pm_instruct_vs_think} presents the parameters for Qwen and Olmo models of different sizes before and after reasoning training phases.
Our results largely confirm that training for mathematical reasoning consistently improves model rationality (including decisiveness) with explicit prospects.
In addition, it significantly increases model predictability (in the sense of goodness of fit for the 4-parameter model as reflected by MSE).
Notably, Qwen2.5-7B-Instruct and Qwen3-30B-Instruct both exhibit near-zero $\beta_{\text{loss}}$ alongside much higher $\beta_{\text{gain}}$, echoing the human pattern of loss-domain randomness observed in Table~\ref{T:explicit}.

\begin{table}[h]
\centering
\begin{minipage}{0.5\textwidth}
\centering
\resizebox{\textwidth}{!}{%
\begin{tabular}{|l|c|c|c|c|c|c|}
\hline
\textbf{Model} 
& $\boldsymbol{\sigma}$ 
& $\boldsymbol{\gamma}$ 
& $\boldsymbol{\beta}_{\text{gain}}$ 
& $\boldsymbol{\beta}_{\text{loss}}$ 
& \textbf{Corr}
& \textbf{MSE}\\
\hline
Qwen2.5-7B-Instruct & 0.84 & 0.75 & 57 & 0.01 & 0.64 & 0.030 \\
Qwen2.5-Math-7B     & 0.85 & 0.82 & 1000 & 403 & 1.00 & 0.001 \\
Qwen3-30B-Instruct  & 0.94 & 0.85 & 56 & 0.01 & 0.57 & 0.041 \\
Qwen3-30B-Thinking  & 0.87 & 0.81 & 726 & 1000 & 0.99 & 0.004 \\
\hline
Olmo-3-7B-Instruct  & 0.88 & 0.81 & 215 & 121 & 0.73 & 0.041 \\
Olmo-3-7B-Think     & 1.33 & 1.16 & 1000 & 1000 & 1.00 & 0.001 \\
Olmo-3-32B-Think    & 1.33 & 1.16 & 1000 & 1000 & 1.00 & 0.000 \\
\hline
\end{tabular}
}
\end{minipage}
\caption{Dual-Beta prospect theory parameters comparing instruction-tuned and reasoning models with explicit prospects.}
\label{tb:pm_instruct_vs_think}
\end{table}

\begin{table}[h]
\centering
\begin{minipage}{0.5\textwidth}
\centering
\resizebox{\textwidth}{!}{%
\begin{tabular}{|l|c|c|c|c|c|c|}
\hline
\textbf{Model} 
& $\boldsymbol{\sigma}$ 
& $\boldsymbol{\gamma}$ 
& $\boldsymbol{\beta}_{\text{gain}}$ 
& $\boldsymbol{\beta}_{\text{loss}}$ 
& \textbf{Corr}
& \textbf{MSE}\\
\hline
Qwen2.5-7B-Instruct & 48 & 0.66 & 0.01 & 2.45 & 0.46 & 0.070 \\
Qwen2.5-Math-7B     & 0.55 & 0.88 & 7 & 0.82 & 0.54 & 0.027 \\
Qwen3-30B-Instruct  & 1.10 & 0.83 & 17 & 2.84 & 0.50 & 0.047 \\
Qwen3-30B-Thinking  & 1.00 & 0.88 & 668 & 585 & 0.94 & 0.025 \\
\hline
Olmo-3-7B-Instruct  & 1.25 & 0.99 & 4.80 & 1.29 & 0.44 & 0.020 \\
Olmo-3-7B-Think     & 0.73 & 0.71 & 1000 & 466 & 1.00 & 0.001 \\
Olmo-3-32B-Think    & 0.95 & 1.07 & 1000 & 235 & 0.93 & 0.032 \\
\hline
\end{tabular}
}
\end{minipage}
\caption{Prospect theory model parameters comparing instruction-tuned and reasoning models with implicit prospects.}
\label{tb:pm_instruct_vs_think_experience}
\end{table}

Table~\ref{tb:pm_instruct_vs_think_experience} presents analogous results with implicit prospects.
Here, we see a similar effect of reasoning-based training with several of the models, although it appears to be somewhat less consistent than with explicit prospects.

In the Supplement (Appendix~\ref{subsec:training_scale}), we report checkpoints from each post-training stage (SFT, DPO, and RLVF) for both the instruction and thinking variants of the open models. As we observe no systematic effects attributable to either DPO or RLVF training, and since \text{Olmo3-7B-Think} and \text{Olmo3-7B-Instruct} share the same base model, one hypothesis can be that the initial SFT stage is the primary source of divergence between a reasoning and a conversational variant of models.

\section{Conclusion}
We study LLM economic behavior through a controlled comparison spanning a diverse model suite and a human baseline. Beyond the canonical context manipulations of option order and outcome framing, we test two levers central to language-mediated decision making—the representation of risky options as explicit prospects versus outcome histories, and explanation prompting—and find that both systematically shift model choices. However, none of the evaluated models are fully human-like in this one-shot risky-choice setting; instead, behavior exhibits a robust two-cluster structure, separating “reasoning” and “conversational” models. Reasoning models are more invariant to contextual perturbations and tend to converge toward an expected payoff-maximizing \ec\ baseline, whereas conversational models remain more context-sensitive and, in some conditions, display more human-like deviations. Our results suggest that LLM economic behavior reflects both model family and the decision interface: how alternatives are represented and responses elicited.

\section{Limitations}
Our analysis exhibits several limitations.
First, our analysis of explanation prompting is incomplete on the human side: we did not elicit all explanation modes from human participants, and we also do not disentangle whether effects depend on the ordering of “decide” versus “explain” (e.g., explanation-before-choice vs choice-before-explanation). Second, in our implicit (sample-based) representation, we aggregate two finite sample lengths (20 and 100 outcomes); because these settings can yield noticeably different behavior, a more systematic treatment of sample size is an important direction for future work. Third, while we vary several factors, fully characterizing their interactions is challenging: joint effects can be non-additive and can induce aggregation artifacts (e.g., Simpson’s paradox), and studying them reliably would likely require substantially more stimuli and a broader set of decision problems than the limited prospect families considered here. Finally, our prospect-theoretic parameterization is intended as a descriptive summary rather than a definitive mechanism: the four-parameter model can suffer from identifiability and boundary-fitting issues in this setting, so fitted parameters should be interpreted cautiously and primarily through robust qualitative comparisons rather than as precise estimates.

Despite these limitations, our work makes important progress on several fronts. The RM--CM 
clustering provides the first systematic behavioral taxonomy of LLMs, the DH gap offers a 
novel window into how LLMs represent uncertain information, and our analysis of training 
effects (mathematical reasoning as the primary differentiator) is one of the first to directly 
test competing explanations for LLM behavioral variation.

\section*{Acknowledgments}
This research was partially supported by the National Science Foundation (IIS-2214141, ITE-2452834), Office of Naval Research (N000142412663), Amazon, and the Foresight Institute.
We are also grateful to Haifeng Xu for discussion and insightful comments.

\bibliography{references}

@inproceedings{mao2025alympics,
  title     = {{ALYMPICS}: {LLM} Agents Meet Game Theory},
  author    = {Mao, Shaoguang and Cai, Yuzhe and Xia, Yan and Wu, Wenshan and Wang, Xun and Wang, Fengyi and Guan, Qiang and Ge, Tao and Wei, Furu},
  booktitle = {Proceedings of the 31st International Conference on Computational Linguistics (COLING 2025)},
  year      = {2025},
  publisher = {Association for Computational Linguistics},
  url       = {https://aclanthology.org/2025.coling-main.193/}
}

@inproceedings{li2024econagent,
  title     = {{Econagent}: Large Language Model-Empowered Agents for Simulating Macroeconomic Activities},
  author    = {Li, Nian and Gao, Chen and Li, Mingyu and Li, Yong and Liao, Qingmin},
  booktitle = {Proceedings of the 62nd Annual Meeting of the Association for Computational Linguistics (Volume 1: Long Papers)},
  year      = {2024}
}

@article{hayes2024relative,
  title={Relative value biases in large language models},
  author={Hayes, William M and Yax, Nicolas and Palminteri, Stefano},
  journal={arXiv preprint arXiv:2401.14530},
  year={2024}
}

@article{argyle2023out,
  title   = {Out of One, Many: Using Language Models to Simulate Human Samples},
  author  = {Argyle, Lisa P. and Busby, Ethan C. and Fulda, Nancy and Gubler, Joshua R. and Rytting, Christopher and Wingate, David},
  journal = {Political Analysis},
  year    = {2023}
}

@inproceedings{coda2024cogbench,
  title={CogBench: a large language model walks into a psychology lab},
  author={Coda-Forno, Julian and Binz, Marcel and Wang, Jane X and Schulz, Eric},
  booktitle={International Conference on Machine Learning},
  year={2024}
}

@article{akata2025playing,
  title   = {Playing Repeated Games with Large Language Models},
  author  = {Akata, Elif and Schulz, Lion and Coda-Forno, Julian and Oh, Seong Joon and Bethge, Matthias and Schulz, Eric},
  journal = {Nature Human Behaviour},
  year    = {2025}
}

@inproceedings{fan2024can,
  title     = {Can Large Language Models Serve as Rational Players in Game Theory? A Systematic Analysis},
  author    = {Fan, Caoyun and Chen, Jindou and Jin, Yaohui and He, Hao},
  booktitle = {Proceedings of the AAAI Conference on Artificial Intelligence},
  year      = {2024}
}

@article{lore2024strategic,
  title   = {Strategic Behavior of Large Language Models and the Role of Game Structure versus Contextual Framing},
  author  = {Lor{\`e}, Nunzio and Heydari, Babak},
  journal = {Scientific Reports},
  year    = {2024}
}

@article{robinson2025framing,
  title         = {Framing the Game: How Context Shapes {LLM} Decision-Making},
  author        = {Robinson, Isaac and Burden, John},
  journal       = {arXiv preprint},
  year          = {2025},
  eprint        = {2503.04840},
  archivePrefix = {arXiv}
}

@article{zhang2024llm,
  title         = {{LLM} as a Mastermind: A Survey of Strategic Reasoning with Large Language Models},
  author        = {Zhang, Yadong and Mao, Shaoguang and Ge, Tao and Wang, Xun and de Wynter, Adrian and Xia, Yan and Wu, Wenshan and Song, Ting and Lan, Man and Wei, Furu},
  journal       = {arXiv preprint},
  year          = {2024},
  eprint        = {2404.01230},
  archivePrefix = {arXiv}
}

@article{mazeika2025utility,
  title         = {Utility Engineering: Analyzing and Controlling Emergent Value Systems in {AI}s},
  author        = {Mazeika, Mantas and Yin, Xuwang and Tamirisa, Rishub and Lim, Jaehyuk and Lee, Bruce W. and Ren, Richard and Phan, Long and Mu, Norman and Khoja, Adam and Zhang, Oliver and others},
  journal       = {arXiv preprint},
  year          = {2025},
  eprint        = {2502.08640},
  archivePrefix = {arXiv}
}

@inproceedings{agnew2024illusion,
  title     = {The Illusion of Artificial Inclusion},
  author    = {Agnew, William and Bergman, A. Stevie and Chien, Jennifer and D{\'\i}az, Mark and El-Sayed, Seliem and Pittman, Jaylen and Mohamed, Shakir and McKee, Kevin R.},
  booktitle = {Proceedings of the 2024 CHI Conference on Human Factors in Computing Systems},
  year      = {2024}
}

@article{myung2000importance,
  title   = {The Importance of Complexity in Model Selection},
  author  = {Myung, In Jae},
  journal = {Journal of Mathematical Psychology},
  year    = {2000}
}

@article{harrison2009expected,
  title   = {Expected Utility Theory and Prospect Theory: One Wedding and a Decent Funeral},
  author  = {Harrison, Glenn W. and Rutstr{\"o}m, E. Elisabet},
  journal = {Experimental Economics},
  year    = {2009}
}

@techreport{bouchouicha2024prospect,
  title       = {Is Prospect Theory Really a Theory of Choice?},
  author      = {Bouchouicha, Ranoua and Oprea, Ryan and Vieider, Ferdinand M. and Wu, Jilong},
  institution = {Working paper},
  year        = {2024}
}

@inproceedings{liu2025evaluating,
  title     = {Evaluating and Aligning Human Economic Risk Preferences in {LLM}s},
  author    = {Liu, Jiaxin and Tang, Yixuan and Yang, Yi and Tam, Kar Yan},
  booktitle = {Proceedings of the 2025 Conference on Empirical Methods in Natural Language Processing},
  year      = {2025}
}

@inproceedings{park2023generative,
  title     = {Generative Agents: Interactive Simulacra of Human Behavior},
  author    = {Park, Joon Sung and {O'Brien}, Joseph and Cai, Carrie Jun and Morris, Meredith Ringel and Liang, Percy and Bernstein, Michael S.},
  booktitle = {Proceedings of the 36th Annual ACM Symposium on User Interface Software and Technology},
  year      = {2023}
}

@article{wilson1991thinking,
  title   = {Thinking Too Much: Introspection Can Reduce the Quality of Preferences and Decisions},
  author  = {Wilson, Timothy D. and Schooler, Jonathan W.},
  journal = {Journal of Personality and Social Psychology},
  year    = {1991}
}

@techreport{horton2023large,
  title       = {Large Language Models as Simulated Economic Agents: What Can We Learn from {Homo Silicus}?},
  author      = {Horton, John J.},
  institution = {National Bureau of Economic Research},
  year        = {2023}
}

@article{chen2023emergence,
  title   = {The Emergence of Economic Rationality of {GPT}},
  author  = {Chen, Yiting and Liu, Tracy Xiao and Shan, You and Zhong, Songfa},
  journal = {Proceedings of the National Academy of Sciences},
  year    = {2023}
}

@article{dillion2023can,
  title   = {Can {AI} Language Models Replace Human Participants?},
  author  = {Dillion, Danica and Tandon, Niket and Gu, Yuling and Gray, Kurt},
  journal = {Trends in Cognitive Sciences},
  year    = {2023}
}

@inproceedings{jiang2025towards,
  title     = {Towards Rationality in Language and Multimodal Agents: A Survey},
  author    = {Jiang, Bowen and Xie, Yangxinyu and Wang, Xiaomeng and Yuan, Yuan and Hao, Zhuoqun and Bai, Xinyi and Su, Weijie J. and Taylor, Camillo Jose and Mallick, Tanwi},
  booktitle = {Proceedings of the 2025 Conference of the Nations of the Americas Chapter of the Association for Computational Linguistics: Human Language Technologies (Volume 1: Long Papers)},
  year      = {2025}
}

@inproceedings{ferrario2022explainability,
  title     = {How Explainability Contributes to Trust in {AI}},
  author    = {Ferrario, Andrea and Loi, Michele},
  booktitle = {Proceedings of the 2022 ACM Conference on Fairness, Accountability, and Transparency},
  year      = {2022}
}

@article{zhao2024explainability,
  title   = {Explainability for Large Language Models: A Survey},
  author  = {Zhao, Haiyan and Chen, Hanjie and Yang, Fan and Liu, Ninghao and Deng, Huiqi and Cai, Hengyi and Wang, Shuaiqiang and Yin, Dawei and Du, Mengnan},
  journal = {ACM Transactions on Intelligent Systems and Technology},
  year    = {2024}
}

@article{festinger1962cognitive,
  title   = {Cognitive Dissonance},
  author  = {Festinger, Leon},
  journal = {Scientific American},
  year    = {1962}
}

@article{shafir1993reason,
  title   = {Reason-Based Choice},
  author  = {Shafir, Eldar and Simonson, Itamar and Tversky, Amos},
  journal = {Cognition},
  year    = {1993},
  doi     = {10.1016/0010-0277(93)90034-S}
}

@techreport{olmo3_2025,
  title       = {Olmo\,3 Technical Report},
  author      = {Ettinger, Allyson and Bertsch, Amanda and Kuehl, Bailey and Graham, David and Heineman, David and Groeneveld, Dirk and Brahman, Faeze and Timbers, Finbarr and Ivison, Hamish and Morrison, Jacob and Poznanski, Jake and Lo, Kyle and Soldaini, Luca and Jordan, Matt and Chen, Mayee and Noukhovitch, Michael and Lambert, Nathan and Walsh, Pete and Dasigi, Pradeep and Berry, Robert and Malik, Saumya and Shah, Saurabh and Geng, Scott and Arora, Shane and Gupta, Shashank and Anderson, Taira and Xiao, Teng and Murray, Tyler and Romero, Tyler and Graf, Victoria and Asai, Akari and Bhagia, Akshita and Wettig, Alex and Liu, Alisa and Rangapur, Aman and Anastasiades, Chloe and Huang, Costa and Schwenk, Dustin and Trivedi, Harsh and Magnusson, Ian and Lochner, Jaron and Liu, Jiacheng and Miranda, Lj and Sap, Maarten and Morgan, Malia and Schmitz, Michael and Guerquin, Michal and Wilson, Michael and Huff, Regan and Le Bras, Ronan and Xin, Rui and Shao, Rulin and Skjonsberg, Sam and Shen, Shannon Zejiang and Li, Shuyue Stella and Wilde, Tucker and Pyatkin, Valentina and Merrill, Will and Chang, Yapei and Gu, Yuling and Zeng, Zhiyuan and Sabharwal, Ashish and Zettlemoyer, Luke and Koh, Pang Wei and Farhadi, Ali and Smith, Noah A. and Hajishirzi, Hannaneh},
  institution = {Allen Institute for AI \& collaborators},
  year        = {2025},
  url         = {https://www.datocms-assets.com/64837/1763662397-1763646865-olmo_3_technical_report-1.pdf}
}

@article{wulff2018meta,
  title   = {A Meta-Analytic Review of Two Modes of Learning and the Description-Experience Gap},
  author  = {Wulff, Dirk U. and Mergenthaler-Canseco, Max and Hertwig, Ralph},
  journal = {Psychological Bulletin},
  year    = {2018}
}

@article{horowitz2025llm,
  title         = {{LLM} Agents Display Human Biases but Exhibit Distinct Learning Patterns},
  author        = {Horowitz, Idan and Plonsky, Ori},
  journal       = {arXiv preprint},
  year          = {2025},
  eprint        = {2503.10248},
  archivePrefix = {arXiv}
}

@article{hertwig2004decisions,
  title   = {Decisions from Experience and the Effect of Rare Events in Risky Choice},
  author  = {Hertwig, Ralph and Barron, Greg and Weber, Elke U. and Erev, Ido},
  journal = {Psychological Science},
  year    = {2004}
}

@article{hertwig2009description,
  title   = {The Description--Experience Gap in Risky Choice},
  author  = {Hertwig, Ralph and Erev, Ido},
  journal = {Trends in Cognitive Sciences},
  year    = {2009}
}

@article{ivanova2023running,
  title         = {Running Cognitive Evaluations on Large Language Models: The Do's and the Don'ts},
  author        = {Ivanova, Anna A.},
  journal       = {arXiv preprint},
  year          = {2023},
  eprint        = {2312.01276},
  archivePrefix = {arXiv}
}

@article{hagendorff2023machine,
  title         = {Machine Psychology},
  author        = {Hagendorff, Thilo and Dasgupta, Ishita and Binz, Marcel and Chan, Stephanie C. Y. and Lampinen, Andrew and Wang, Jane X. and Akata, Zeynep and Schulz, Eric},
  journal       = {arXiv preprint},
  year          = {2023},
  eprint        = {2303.13988},
  archivePrefix = {arXiv}
}

@article{ross2024llm,
  title         = {{LLM} Economicus? Mapping the Behavioral Biases of {LLM}s via Utility Theory},
  author        = {Ross, Jillian and Kim, Yoon and Lo, Andrew W.},
  journal       = {arXiv preprint},
  year          = {2024},
  eprint        = {2408.02784},
  archivePrefix = {arXiv}
}

@article{wang2025prospect,
  title         = {Prospect Theory Fails for {LLM}s: Revealing Instability of Decision-Making under Epistemic Uncertainty},
  author        = {Wang, Rui and Lin, Qihan and Liu, Jiayu and Zong, Qing and Zheng, Tianshi and Wang, Weiqi and Song, Yangqiu},
  journal       = {arXiv preprint},
  year          = {2025},
  eprint        = {2508.08992},
  archivePrefix = {arXiv}
}

@article{connolly2006regret,
  title   = {Regret in Economic and Psychological Theories of Choice},
  author  = {Connolly, Terry and Butler, David},
  journal = {Journal of Behavioral Decision Making},
  year    = {2006}
}

@article{Kahneman1979PT,
  title   = {Prospect Theory: An Analysis of Decision under Risk},
  author  = {Kahneman, Daniel and Tversky, Amos},
  journal = {Econometrica},
  year    = {1979}
}

@incollection{kahneman2013prospect,
  title     = {Prospect Theory: An Analysis of Decision under Risk},
  author    = {Kahneman, Daniel and Tversky, Amos},
  booktitle = {Handbook of the Fundamentals of Financial Decision Making: Part I},
  year      = {2013},
  publisher = {World Scientific}
}

@article{payne2025analysis,
  title         = {An Analysis of {AI} Decision under Risk: Prospect Theory Emerges in Large Language Models},
  author        = {Payne, Kenneth},
  journal       = {arXiv preprint},
  year          = {2025},
  eprint        = {2508.00902},
  archivePrefix = {arXiv}
}

@article{jia2024decision,
  title   = {Decision-Making Behavior Evaluation Framework for {LLM}s under Uncertain Context},
  author  = {Jia, Jingru Jessica and Yuan, Zehua and Pan, Junhao and McNamara, Paul and Chen, Deming},
  journal = {Advances in Neural Information Processing Systems},
  year    = {2024}
}

@article{zhou2025shared,
  title   = {Shared Imagination: {LLM}s Hallucinate Alike},
  author  = {Zhou, Yilun and Xiong, Caiming and Savarese, Silvio and Wu, Chien-Sheng},
  journal = {Transactions on Machine Learning Research},
  year    = {2025}
}

@article{smith2025comprehensive,
  title         = {A Comprehensive Analysis of Large Language Model Outputs: Similarity, Diversity, and Bias},
  author        = {Smith, Brandon and Bouadjenek, Mohamed Reda and Kheya, Tahsin Alamgir and Dawson, Phillip and Aryal, Sunil},
  journal       = {arXiv preprint},
  year          = {2025},
  eprint        = {2505.09056},
  archivePrefix = {arXiv}
}

@article{peterson2021using,
  title   = {Using Large-Scale Experiments and Machine Learning to Discover Theories of Human Decision-Making},
  author  = {Peterson, Joshua C. and Bourgin, David D. and Agrawal, Mayank and Reichman, Daniel and Griffiths, Thomas L.},
  journal = {Science},
  year    = {2021}
}

@article{binz2023using,
  title={Using cognitive psychology to understand GPT-3},
  author={Binz, Marcel and Schulz, Eric},
  journal={Proceedings of the National Academy of Sciences},
  volume={120},
  number={6},
  pages={e2218523120},
  year={2023},
}

@article{benary2023leveraging,
  title   = {Leveraging Large Language Models for Decision Support in Personalized Oncology},
  author  = {Benary, Manuela and Wang, Xing David and Schmidt, Max and Soll, Dominik and Hilfenhaus, Georg and Nassir, Mani and Sigler, Christian and Kn{\"o}dler, Maren and Keller, Ulrich and Beule, Dieter and others},
  journal = {JAMA Network Open},
  year    = {2023}
}

@inproceedings{vrdoljak2025review,
  title     = {A Review of Large Language Models in Medical Education, Clinical Decision Support, and Healthcare Administration},
  author    = {Vrdoljak, Josip and Boban, Zvonimir and Vilovi{\'c}, Marino and Kumri{\'c}, Marko and Bo{\v z}i{\'c}, Jo{\v s}ko},
  booktitle = {Healthcare},
  year      = {2025}
}

@article{moncada2025agentic,
  title   = {Agentic Workflows for Improving Large Language Model Reasoning in Robotic Object-Centered Planning},
  author  = {Moncada-Ramirez, Jesus and Matez-Bandera, Jose-Luis and Gonzalez-Jimenez, Javier and Ruiz-Sarmiento, Jose-Raul},
  journal = {Robotics},
  year    = {2025}
}

@article{webb2025brain,
  title   = {A Brain-Inspired Agentic Architecture to Improve Planning with {LLM}s},
  author  = {Webb, Taylor and Mondal, Shanka Subhra and Momennejad, Ida},
  journal = {Nature Communications},
  year    = {2025}
}

@article{parkes2015economic,
  title   = {Economic Reasoning and Artificial Intelligence},
  author  = {Parkes, David C. and Wellman, Michael P.},
  journal = {Science},
  year    = {2015}
}

@article{du2024understanding,
  title   = {Understanding Emergent Abilities of Language Models from the Loss Perspective},
  author  = {Du, Zhengxiao and Zeng, Aohan and Dong, Yuxiao and Tang, Jie},
  journal = {Advances in Neural Information Processing Systems},
  year    = {2024}
}

@article{guo2025deepseek,
  title         = {Deepseek-R1: Incentivizing Reasoning Capability in {LLM}s via Reinforcement Learning},
  author        = {Guo, Daya and Yang, Dejian and Zhang, Haowei and Song, Junxiao and Zhang, Ruoyu and Xu, Runxin and Zhu, Qihao and Ma, Shirong and Wang, Peiyi and Bi, Xiao and others},
  journal       = {arXiv preprint},
  year          = {2025},
  eprint        = {2501.12948},
  archivePrefix = {arXiv}
}

@article{ouyang2022training,
  title   = {Training Language Models to Follow Instructions with Human Feedback},
  author  = {Ouyang, Long and Wu, Jeffrey and Jiang, Xu and Almeida, Diogo and Wainwright, Carroll and Mishkin, Pamela and Zhang, Chong and Agarwal, Sandhini and Slama, Katarina and Ray, Alex and others},
  journal = {Advances in Neural Information Processing Systems},
  year    = {2022}
}
\clearpage

\appendix

\section*{Appendix Table of Contents}

\noindent
\textbf{\ref{sec:app_related_work} Additional Related Works}
\vspace{0.6em}

\noindent
\textbf{\ref{sec:app_design} Additional Study Design Details} \par
\vspace{0.2em}
\noindent \hspace{1.5em} \ref{subsec:llmslist} LLMs Used \par
\noindent \hspace{1.5em} \ref{subsec:model} Behavior Models \par
\noindent \hspace{3em} \ref{subsubsec:pt_model} Prospect Theory \par
\noindent \hspace{3em} \ref{subsubsec:ra_model} Regret Aversion \par
\noindent \hspace{3em} \ref{subsubsec:model_fitting} Model Fitting \par
\noindent \hspace{1.5em} \ref{subsec:prompts} Prompts \par
\noindent \hspace{1.5em} \ref{subsec:base_prospects} Prospects
\vspace{0.6em}

\noindent
\textbf{\ref{sec:app_results} Additional Experimental Results} \par
\vspace{0.2em}
\noindent \hspace{1.5em} \ref{subsec:heatmaps} Correlation Heatmaps \par
\noindent \hspace{1.5em} \ref{subsec:consistency} Consistency and Decisiveness \par
\noindent \hspace{1.5em} \ref{subsec:sample_size} Sample Size Effects for Implicit Prospect \par
\noindent \hspace{1.5em} \ref{subsec:training_scale} Impact of Training and Scale for Open Models \par
\noindent \hspace{1.5em} \ref{subsec:ci_results} Confidence Intervals \par
\noindent \hspace{1.5em} \ref{subsec:baseline_results} Ablation and Alternative Model Results \par
\noindent \hspace{3em} \ref{subsubsec:restricted_pt} Restricted Prospect Theory Models \par
\noindent \hspace{3em} \ref{subsubsec:ra} Regret Aversion \par
\noindent \hspace{1.5em} \ref{subsec:old_pt_results} Standard Prospect Theory Results \par
\vspace{0.6em}

\noindent
\textbf{\ref{sec:human_instructions} Human Instructions}

\section*{Overview}
This appendix provides supplementary information to support the main analysis.

\section{Additional Related Works}
\label{sec:app_related_work}
\textbf{Behavioral models for economic behavior.}
In the main body, we summarize different agents' behavior using an augmented prospect-theoretic model.  Beyond prospect theory~\cite{Kahneman1979PT}, a broad family of models has been proposed for risky choice, including regret--rejoicing accounts \cite{connolly2006regret}, mixture models \cite{harrison2009expected}, and more flexible neural approaches \cite{peterson2021using}. These alternatives often trade off interpretability against predictive flexibility, and model comparison in practice is constrained by identifiability and data requirements \cite{myung2000importance}. In light of ongoing debates about the descriptive adequacy of prospect theory for human choice \cite{bouchouicha2024prospect}, we treat our parameter estimates as a compact behavioral summary rather than a definitive mechanism; developing and validating richer models that better capture LLM-specific decision artifacts is an important direction for future work.

\noindent \textbf{LLM agents and LLMs as human simulators.}
LLM-based agents are increasingly used in interactive systems and simulations, sometimes explicitly as proxies for human participants (e.g., ``silicon samples'' and generative agents) \cite{argyle2023out,park2023generative}. Closest in spirit to our setting are economically-motivated agent frameworks such as Homo Silicus~\cite{horton2023large} and EconAgent \cite{li2024econagent}. However, both our results and prior work~\cite{agnew2024illusion,horowitz2025llm} suggest systematic gaps between LLM and human behavior, implying that LLM-based human simulation should be interpreted cautiously, especially when used for social-scientific inference or high-stakes product decisions.

\noindent \textbf{Other types of LLM decision making.}
Beyond one-shot risky choice, an active line of work evaluates LLM decision making in strategic interaction. Prior studies examine classical game settings, from matrix games \cite{fan2024can,akata2025playing,mao2025alympics} to richer multi-agent environments (see \cite{zhang2024llm} for a survey), and generally find that strong instruction-following and reasoning ability does not reliably translate into game-theoretic rationality. At the same time, this area would benefit from larger-scale and more systematic protocols; recent work has begun to probe strategic behavior under controlled perturbations, including contextual and framing-style effects \cite{lore2024strategic,robinson2025framing}.

\section{Additional Study Design Details}
\label{sec:app_design}

\subsection{LLMs Used}
\label{subsec:llmslist}

As mentioned in the main body, our model suite is designed to (i) cover widely used frontier black-box systems from major providers and (ii) enable controlled comparisons using open-weight models spanning different sizes and training stages.
Specifically, we include 7 black-box models (\textit{Claude Haiku-4.5, GPT-4.1, GPT-5.1, Gemini-2.5-Flash, Gemini-2.5-Pro, DeepSeek Chat, DeepSeek R1}),
7 open-weight models (\textit{
Qwen2.5-7B-Instruct, Qwen2.5-Math-7B-Instruct, Qwen2.5-30B-instruct, Qwen2.5-30B-math-instruct, Olmo3-7B-think, Olmo3-32B-think, Olmo3-7B-instruct}),
and 6 intermediate checkpoints within the OLMo3 family (\textit{Olmo3-7B-think-sft, Olmo3-7B-think-dpo, Olmo3-7B-instruct-sft, Olmo3-7B-instruct-dpo, Olmo3-32B-think-sft, Olmo3-32B-think-dpo}).
This set spans multiple teams (Anthropic, OpenAI, Google, DeepSeek, Meta, Qwen, Allen-AI), scales (7B--32B for open-weight models), and post-training stages (instruction and reasoning SFT $\rightarrow$ DPO $\rightarrow$ RLVF).

\subsection{Behavior Models}
\label{subsec:model}

\subsubsection{Prospect Theory}
\label{subsubsec:pt_model}

Our parametric model builds upon the framework adopted by \citet{wang2025prospect}, which follows a standard prospect theory model from prior literature. In this framework, decision-making is modeled via two distinct components: a \textit{value function} $v(x)$ that maps objective outcomes to subjective utility, and a \textit{probability weighting function} $w(p)$ that accounts for the non-linear perception of probabilities, combined with a stochastic choice rule. We consider two parameterizations: a \textit{standard} formulation with a loss aversion coefficient $\lambda$ and a single decisiveness parameter $\beta$, and a \textit{dual-beta} (generalized power value model~\cite{peterson2021using}) specification that replaces these with domain-specific decisiveness parameters $\beta_{\text{gain}}$ and $\beta_{\text{loss}}$. We describe the dual-beta specification below; the standard formulation differs only in the value function and choice rule, as detailed at the end of this section.

Subjective utility is formalized via the \textbf{value function}, $v(x)$. In the dual-beta specification, we utilize a symmetric power function:

\begin{equation}
v(x) = 
\begin{cases}
x^\sigma & \text{for } x\geq0 \\
-(-x)^\sigma & \text{for } x<0.
\end{cases}
\end{equation}

This functional form is modulated by the curvature parameter $\sigma$ (risk preference), governing the marginal sensitivity to payoff magnitude.

To capture the non-linear distortion of objective probabilities $p$ into subjective decision weights, we utilize the canonical \textbf{probability weighting function} introduced by \citet{Kahneman1979PT}:

\begin{equation}
    w(p) = \frac{p^\gamma}{\left(p^\gamma + (1-p)^\gamma \right)^{1/\gamma}},
\end{equation}
where the parameter $\gamma$ modulates the curvature of the weighting function.

The aggregate utility for a binary prospect in the form $P = (x, p; y, q)$ is defined as follows:
\begin{equation}
u(P) =
\begin{cases}
v(y) + w(p)(v(x) - v(y)) & \mkern-15mu [1]\\
w(p)v(x) + w(q)v(y) & \mkern-15mu [2],
\end{cases}
\end{equation}
where $[1]$ is "$\text{if } x > y > 0 \text{ or } x < y < 0$" and $[2]$ is "$\text{if } x < 0 < y$".

Finally, we define the predicted probability of choosing option A for each lottery using a logistic choice rule with \textit{domain-dependent decisiveness}. The probability is given by:
\begin{equation}
P(\text{choose A}) = \frac{1}{1 + e^{-\Delta}},
\end{equation}
where $\Delta = \beta_{context} \cdot (u(\text{A}) - u(\text{B}))$. The precision parameter $\beta_{context}$ varies depending on the outcome domain:

\begin{equation}
\beta_{context} = 
\begin{cases}
\beta_{gain} & \text{if } x_A, y_A, x_B, y_B \geq 0 \\
\beta_{loss} & \text{otherwise}.
\end{cases}
\end{equation}

In this framework, the standard concept of loss aversion is implicitly captured by the ratio between $\beta_{loss}$ and $\beta_{gain}$. Specifically, a ratio $\beta_{loss} / \beta_{gain} > 1$ implies that the model is more sensitive to utility differences (steeper value slope) in the loss domain than in the gain domain, serving as a direct proxy for the traditional loss aversion parameter $\lambda$. We minimize MSE to estimate the four learnable parameters: $\sigma, \gamma, \beta_{gain}, \text{ and } \beta_{loss}$.

\paragraph{Standard Formulation.}
The standard prospect theory parameterization replaces the dual-beta choice rule with a single decisiveness parameter $\beta$ and introduces a loss aversion coefficient $\lambda$ into the value function:
\begin{equation}
v(x) = 
\begin{cases}
x^\sigma & \text{for } x\geq0 \\
-\lambda(-x)^\sigma & \text{for } x<0,
\end{cases}
\end{equation}
with the choice probability given by $P(\text{choose A}) = \sigma({\beta \cdot (u(A)-u(B))})$, where $\sigma(\cdot)$ is the logistic function. This yields four parameters: $\sigma$, $\lambda$, $\gamma$, and $\beta$. However, in pure-loss prospects---where all payoffs are non-positive---$\lambda$ factors out of every value term, so the choice probability depends only on the product $\beta \cdot \lambda$ rather than on each parameter individually. The two are therefore not jointly identifiable in the loss domain. We adopt the dual-beta specification as our primary model to avoid this issue; results for the standard formulation are provided in Appendix~\ref{subsec:old_pt_results} and yield qualitatively similar conclusions.

\subsubsection{Regret Aversion}
\label{subsubsec:ra_model}

As an alternative to the reference-dependent valuation of Prospect Theory, we also consider a regret-based model. In this framework, the utility of an option is derived not from independent evaluation, but from a direct pairwise comparison of outcomes. 

We adopt a parameterized regret function where the subjective evaluation $Q$ depends on the difference $\delta = x_A - x_B$ between the payoffs of the two options. The evaluation function is defined as:

\begin{equation}
    Q(\delta) = \delta + \kappa \cdot \text{sgn}(\delta) \cdot |\delta|^\alpha,
\end{equation}

\noindent where $\kappa$ weights the non-linear regret (or rejoicing) component, and $\alpha$ controls the curvature or sensitivity of this term relative to the payoff difference.

The total expected regret-adjusted value for a prospect $A$, denoted $R_A$, is computed by aggregating these evaluations over the distribution of outcomes:

\begin{equation}
    R_A = \sum_{i} p_i Q(x_{A,i} - x_{B,i}),
\end{equation}

\noindent where the summation is taken over the corresponding outcome pairs of the two prospects. The value $R_B$ is computed analogously. The final probability of selecting Option A is determined via a logistic function of the difference in these values:

\begin{equation}
    P(\text{Choose A}) = \frac{1}{1 + e^{-\lambda_{\text{reg}} (R_A - R_B)}},
\end{equation}

\noindent where $\lambda_{\text{reg}}$ acts as a decisiveness parameter (analogous to $\beta$ in our Prospect Theory specification). This model yields three learnable parameters: $\kappa$, $\alpha$, and $\lambda_{\text{reg}}$.

\subsubsection{Model Fitting}
\label{subsubsec:model_fitting}

We fit our models by minimizing the Mean Squared Error (MSE) between the model's predicted selection probabilities and the observed empirical choice rates. For a dataset of $N$ trials, the objective function is:

\begin{equation}
    \min_{\theta} \frac{1}{N} \sum_{j=1}^{N} \left( p_{\text{obs}}^{(j)} - p_{\text{pred}}^{(j)}(\theta) \right)^2
\end{equation}

\noindent We optimize this objective using the L-BFGS-B algorithm which allows for bound constraints on parameters. This ensures that parameters remain within theoretically valid ranges.

\paragraph{Model Variants and Starting Points.}
To ensure robustness and avoid local minima, we implement a multi-start strategy for every fit ($n_{\text{starts}}=20$). We fit five model variants: the dual-beta PT model used in the main analysis, three variants of the standard PT parameterization to isolate different behavioral components, and a regret-based alternative. For each variant, we utilize a combination of fixed baselines and random initializations:

\begin{itemize}
    \item \textbf{Dual-Beta PT ($\sigma, \gamma, \beta_{\text{gain}}, \beta_{\text{loss}}$).} This is the primary model reported in the main text.
    \begin{itemize}
        \item \textit{Starts:} Fixed starts at $[1, 1, 1, 1]$, $[1, 1, 1000, 1]$, $[1, 1, 1, 1000]$, and $[1, 1, 1000, 1000]$, plus random samples with $\sigma, \gamma \sim U(0.01, 3)$ and $\beta_{\text{gain}}, \beta_{\text{loss}} \sim U(0.01, 100)$.
    \end{itemize}
    
    \item \textbf{Model 1: Beta-Only PT ($\beta$).} This model assumes Expected Utility (fixing $\sigma=\lambda=\gamma=1$) and fits only the decisiveness parameter $\beta$. 
    \begin{itemize}
        \item \textit{Starts:} Fixed starts at $\beta=1.0$ (neutral) and $\beta=1000.0$ (deterministic), plus random samples $\beta \sim U(0.01, 100)$.
    \end{itemize}

    \item \textbf{Model 2: Shape-Only PT ($\sigma, \lambda, \gamma$).} This model fits the risk and loss parameters while fixing noise/decisiveness at $\beta=1$.
    \begin{itemize}
        \item \textit{Starts:} Fixed start at $[1, 1, 1]$ (risk neutral), plus random samples for all parameters $\sim U(0.01, 3)$.
    \end{itemize}

    \item \textbf{Model 3: Full PT ($\sigma, \lambda, \gamma, \beta$).} This model fits all four parameters simultaneously. To improve convergence, we employ a "warm start" strategy using the best fits from the simpler models.
    \begin{itemize}
        \item \textit{Starts:} Fixed starts at $[1, 1, 1, 1]$ and $[1, 1, 1, 1000]$. Crucially, we also include the composed solution from Models 1 and 2 ($[\sigma_{m2}, \gamma_{m2}, \lambda_{m2}, \beta_{m1}]$) as a starting point, alongside random samples with $\sigma, \lambda, \gamma \sim U(0.01, 3)$ and $\beta \sim U(0.01, 100)$.
    \end{itemize}

    \item \textbf{Model 4: Regret Theory ($\lambda_{\text{reg}}, \kappa, \alpha$).} This fits the three parameters of the regret model.
    \begin{itemize}
        \item \textit{Starts:} Fixed start at $[1, 1, 1.5]$ (standard regret), plus random samples for $\lambda_{\text{reg}} \sim U(0.01, 1000)$ and shape parameters $\kappa, \alpha \sim U(0, 1000)$.
    \end{itemize}
\end{itemize}

\paragraph{Robustness Checks.}
For every fit, the parameter set yielding the lowest MSE across all initialization attempts is selected as the final estimate. Parameters are bounded to $[0.01, 1000]$ to prevent numerical instability, and payoff magnitudes are normalized by the maximum absolute payoff in the batch prior to fitting.

\paragraph{Uncertainty Estimation.}
To assess the reliability of our parameter estimates, we calculated 95\% confidence intervals using a parametric bootstrap procedure.
First, we computed the predicted choice probabilities $\hat{p}_i$ for each prospect pair $i$ using the best-fitting parameters $\hat{\theta}$ derived from the original data.
We then generated $B=1000$ synthetic datasets; for each dataset, the number of times Option A was chosen for prospect $i$ was simulated from a Binomial distribution $\text{Bin}(n=20, p=\hat{p}_i)$, reflecting the approximately 20 valid trials per condition in our data collection.
The model was re-fitted to each synthetic dataset to obtain a distribution of bootstrap estimates $\{\hat{\theta}^*_1, \dots, \hat{\theta}^*_B\}$.
The 95\% confidence intervals were determined using the percentile method, defined by the $2.5^{th}$ and $97.5^{th}$ percentiles of the bootstrap distribution.

\subsection{Prompts and Prompts Sensitivity Tests}
\label{subsec:prompts}

\noindent Each prompt is the concatenation of three parts:
(i) a \emph{context block} (explicit vs.\ implicit prospect),
(ii) an \emph{explanation instruction} (one of three styles),
and (iii) the \emph{choice block} presenting Options A and B.
\noindent{The template used for all conditions is the following: }
\begin{quote}\ttfamily
\{CONTEXT\_BLOCK\}

\{EXPLANATION\_INSTRUCTION\}

Option A: \{A\_TEXT\}

Option B: \{B\_TEXT\}
\end{quote}

\noindent 
Inside the context block, it's one of the two:

\begin{description}
  \item[\texttt{Explicit prospect:}]
  You will be provided with two options with different payoffs and uncertainties.

  \item[\texttt{Implicit prospect:}]
  You will be provided with two options that are histories of past payoffs.
\end{description}

\noindent Inside the explanation-instruction block, it's one of the three: 

\begin{description}
  \item[\texttt{no\_explanation}]
  Respond your choice with `A' or `B' only.

  \item[\texttt{one\_sentence\_explanation}]
  Respond your choice with `A' or `B' plus one brief sentence for explanation.

  \item[\texttt{math\_explanation}]
  Respond your choice with `A' or `B' plus a brief mathematical explanation.
\end{description}

Additionally, we conducted prompt sensitivity analysis using three variations: "You will be provided with...", "Here are...", "Please consider...". We collected data from 4 RMs (gpt-5.1, gemini-2.5-pro, Olmo3-7B-think,Qwen2.5-Math-instruct) and 4 CMs (gpt-4.1, gemini-2.5-flash, Olmo3-7B-instruct,Qwen2.5-instruct). Overall, we found that the RM-CM gap is robust across prompt variations. In particular, the range of RM–RM correlation is (0.82 to 0.99) with mean 0.91, whereas the range of RM–CM correlation is (0.21 to 0.78) with mean of 0.31.

\subsection{Prospects}
\label{subsec:base_prospects}

\noindent We use three base prospects as a small but structured set for probing the macro-level trends of different LLMs' risky choice behavior. 
Together they vary (i) the \emph{scale} of outcomes (single-digit, hundreds, thousands), (ii) the \emph{probability regime} (low, medium, high; including a sure loss), and (iii) how \emph{close} the options are in expected value (near-ties vs.\ clearer separations). 
This design lets us test whether model behavior is stable across salient payoff magnitudes and uncertainty levels while keeping the task minimal.

In the \texttt{explicit} condition, each pair is presented directly in terms of outcomes and probabilities. In the \texttt{implicit} condition, we present payoff histories generated by sampling $n=\{20,100\}$ returns from the same underlying distributions using four seeds. 
In total, each model answers 6 \texttt{explicit} prospect questions (3 base prospects $\times$ 2 frames) and 48 \texttt{implicit} prospect questions (3 base prospects $\times$ 2 frames $\times$ (4 histories with $n{=}20$ + 4 histories with $n{=}100$)). Crossing these with two option orderings and three explanation instructions yields $(6+48)\times 2 \times 3 = 324$ questions per model.

\begin{itemize}

  \item \textbf{Base Prospect 1 .}\\
  Option A: (Lose) 100 with probability 0.33; otherwise 0.\\
  Option B: (Lose) 96 with probability 0.34; otherwise 0.

  \item \textbf{Base  Prospect 2 .}\\
  Option A: (Lose) 5000 with probability 0.80; otherwise 0.\\
  Option B: (Lose) 3500 with certainty.

  \item \textbf{Base  Prospect 3 .}\\
  Option A: (Lose) 7 with probability 0.10; otherwise 0.\\
  Option B: (Lose) 4 with probability 0.20; otherwise 0.
\end{itemize}

\section{Additional Experimental Results}
\label{sec:app_results}

\subsection{Correlations Heatmaps}
\label{subsec:heatmaps}

\begin{figure*}[h!]
    \centering
    \includegraphics[width=\linewidth]{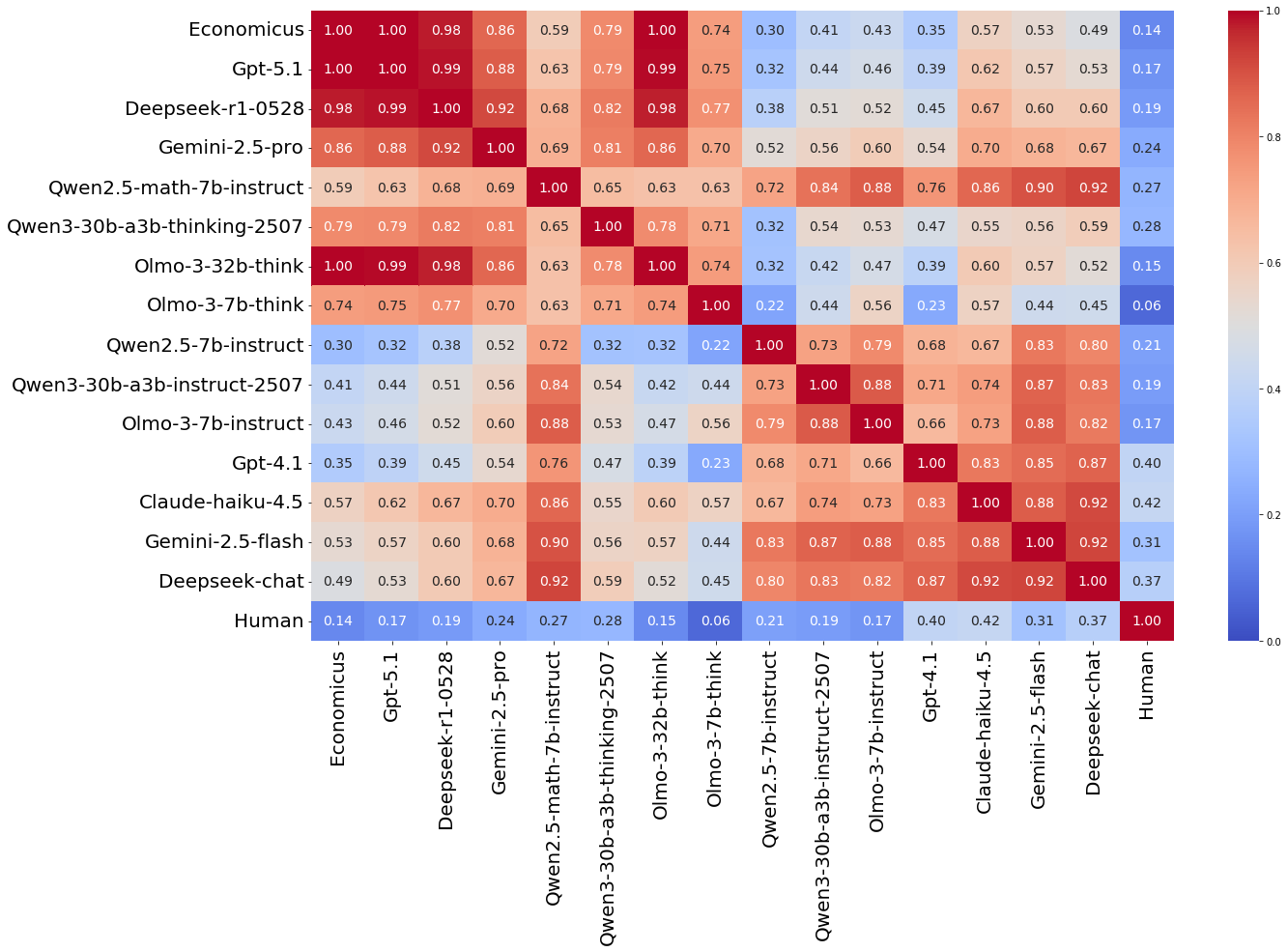}
    \caption{Correlation matrix involving (1) LLMs, (2) \ec, and (3) human responses to all questions.}
       \label{fig:heatmap_full}
\end{figure*}

\begin{figure*}[h!]
    \centering
    \includegraphics[width=\linewidth]{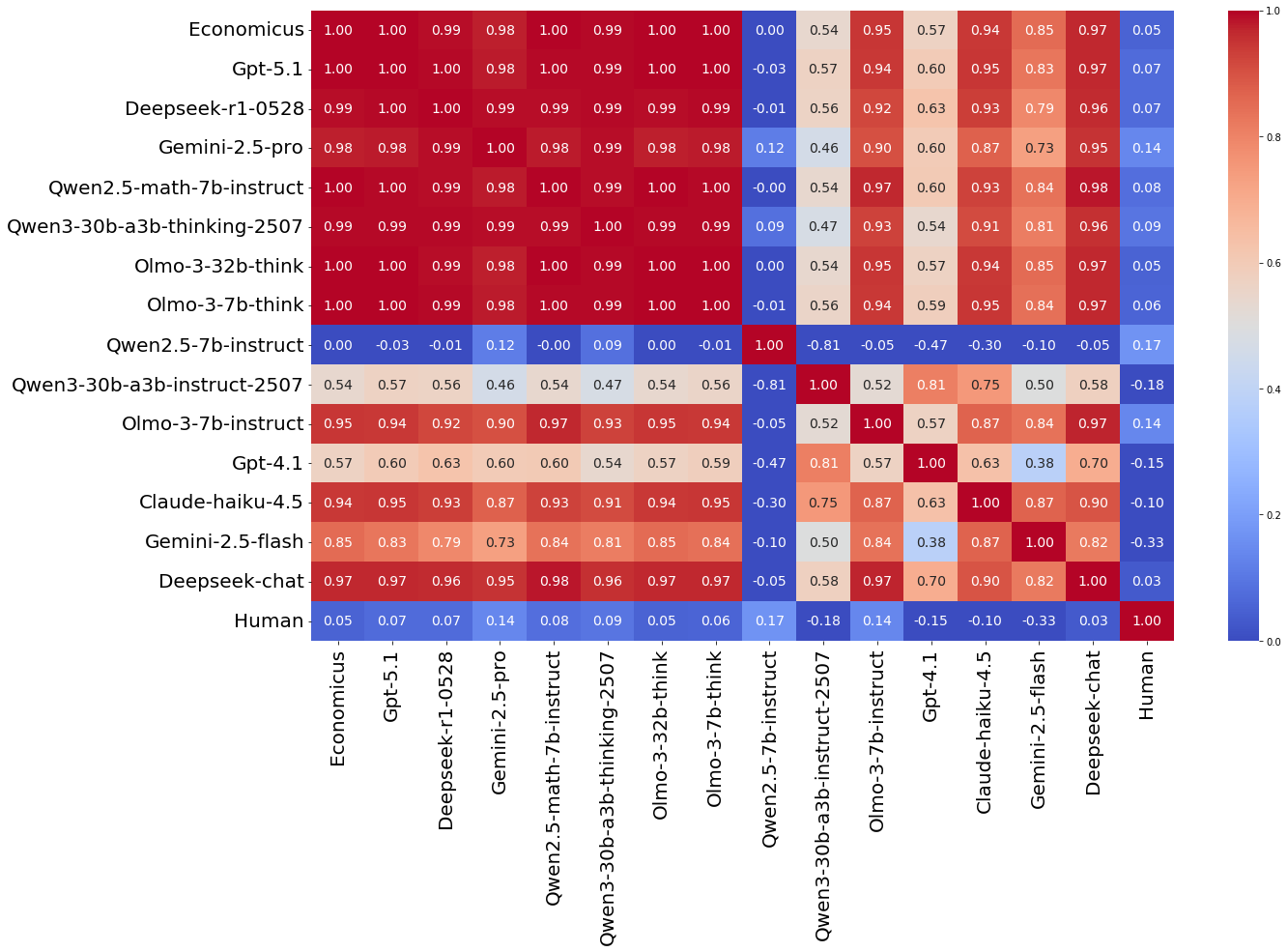}
    \caption{Correlation matrix involving (1) LLMs, (2) \ec, and (3) human responses to all questions with explicit prospects.}
       \label{fig:heatmap_full_e}
\end{figure*}

\begin{figure*}[h!]
    \centering
    \includegraphics[width=\linewidth]{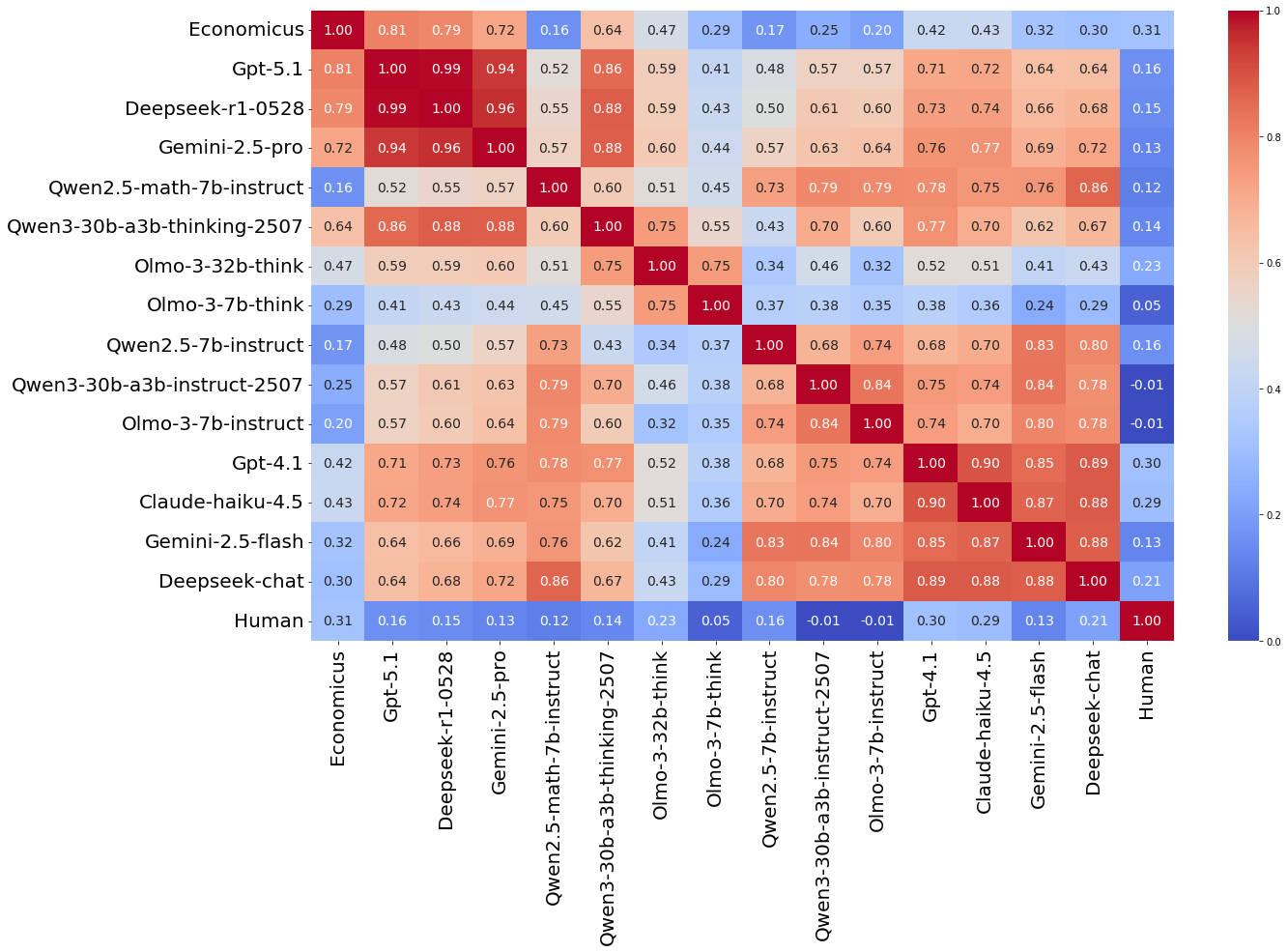}
    \caption{Correlation matrix involving (1) LLMs, (2) \ec, and (3) human responses to all questions with implicit prospects.}
       \label{fig:heatmap_full_i}
\end{figure*}

Examining the full heatmap in Figure~\ref{fig:heatmap_full}, we observe that the open models can also be grouped into two clusters: RMs (Qwen2.5-math-7B-instruct, Qwen3-30B-think, Olmo3-32B-think, Olmo3-7B-think) and CMs (Qwen2.5-7B-instruct, Qwen3-30B-instruct, Olmo3-7B-instruct). Notably, for explicit prospects, the RMs exhibit high mutual correlation at least 0.98, as well as strong correlation with \ec, which manifests as the prominent red block in the top-left corner of the heatmap. In contrast, for implicit prospects, the CMs form a more coherent cluster in the bottom-right corner, with correlations exceeding 0.68 regardless of model size. In both settings, however, the models display relatively weak similarity to human performance.

\subsection{Consistency and Decisiveness}
\label{subsec:consistency}

\begin{figure*}[h!]
    \centering
    \includegraphics[width=0.7
    \linewidth]{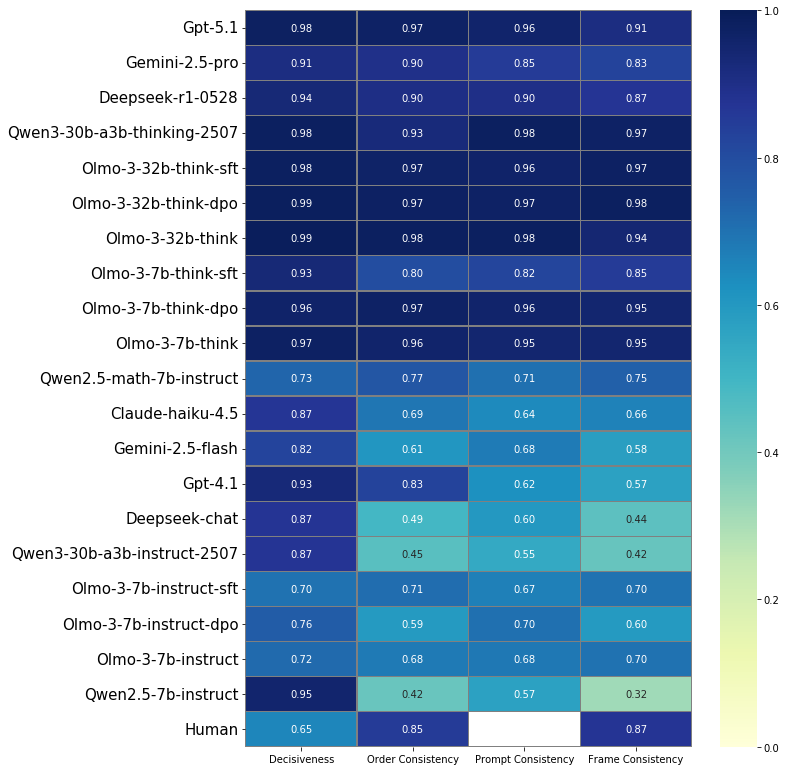}
    \caption{Consistency and decisiveness heatmap for all models and the human subjects.}
    \label{fig:c_d_full}
\end{figure*}
Figure~\ref{fig:c_d_full} shows that the patterns observed in frontier models also generalize to open models: RMs generally exhibit higher decisiveness and consistency. Interestingly, humans appear more decisive but less consistent when transitioning from explicit to implicit prospects, with changes of at least 6\%, a trend not observed in any of the models.

 \begin{figure*}[h!]
     \centering
     \includegraphics[width=0.7\linewidth]{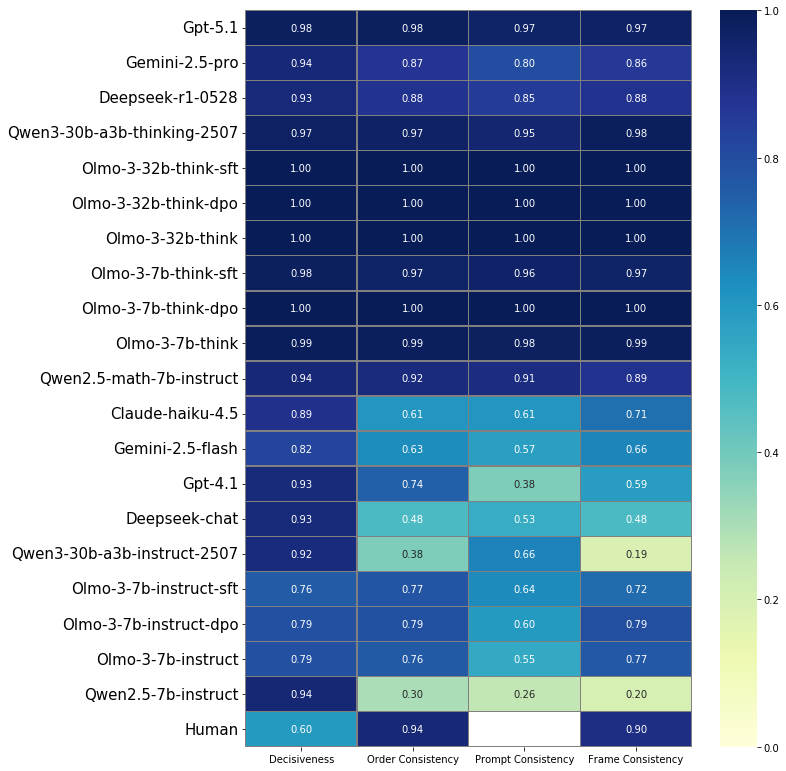}
     \caption{Consistency and decisiveness heatmap for all models and the human subjects restricted to the explicit prospect setting.}
     \label{fig:c_d_explicit}
 \end{figure*}

 \begin{figure*}[h!]
     \centering
     \includegraphics[width=0.7\linewidth]{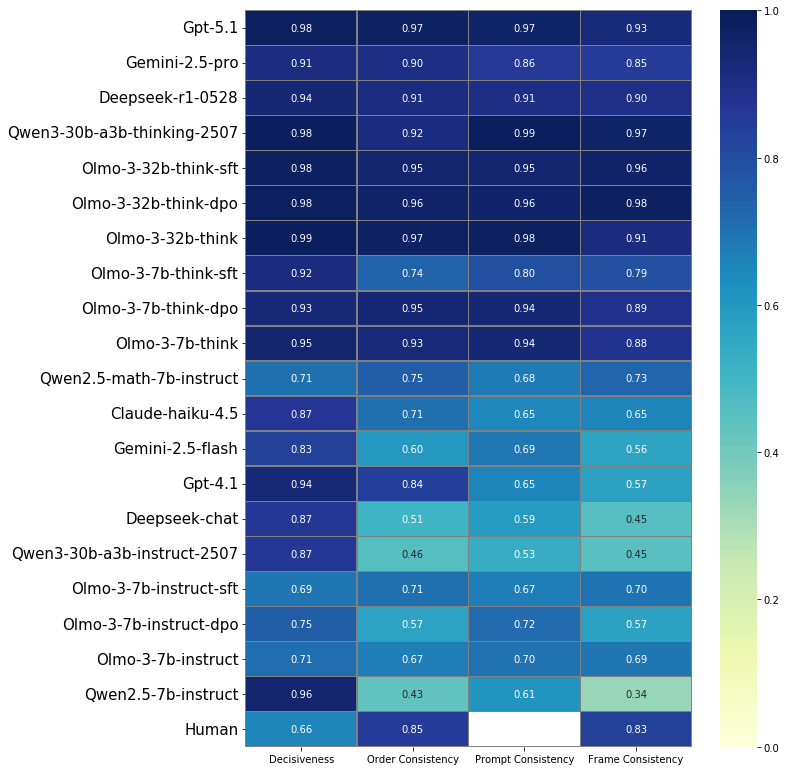}
     \caption{Consistency and decisiveness heatmap for all models and the human subjects restricted to the implicit prospect setting.}
     \label{fig:c_d_implicit}
 \end{figure*}

\subsection{Sample Size Effects for Implicit Prospect}
\label{subsec:sample_size}

\begin{figure}[h]
    \centering
    \includegraphics[width=\linewidth]{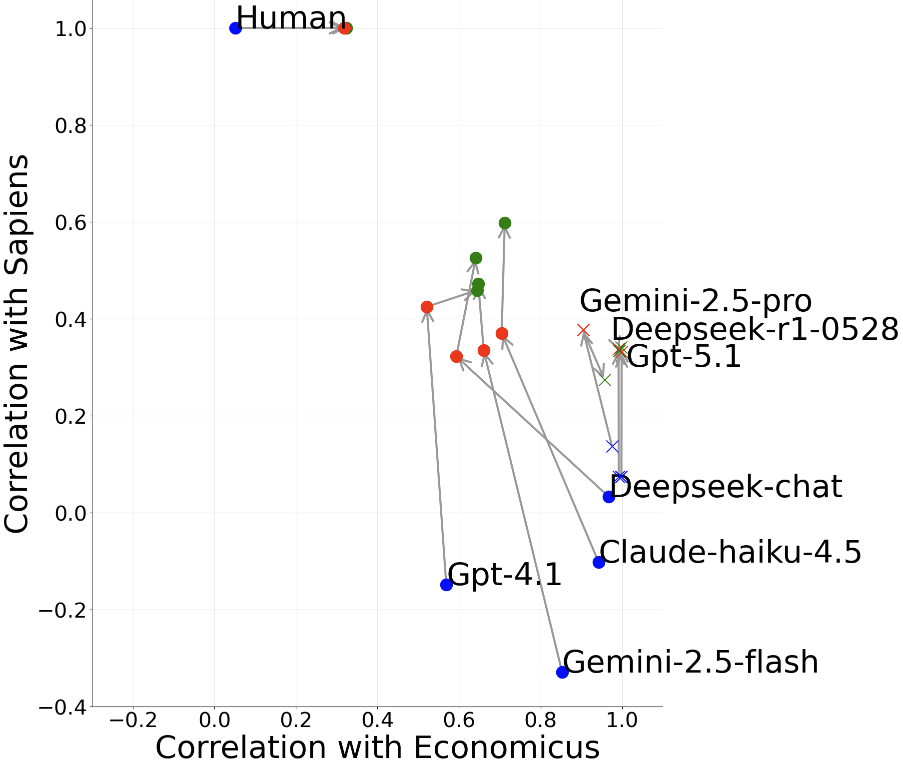}
    
\includegraphics[width=\linewidth]{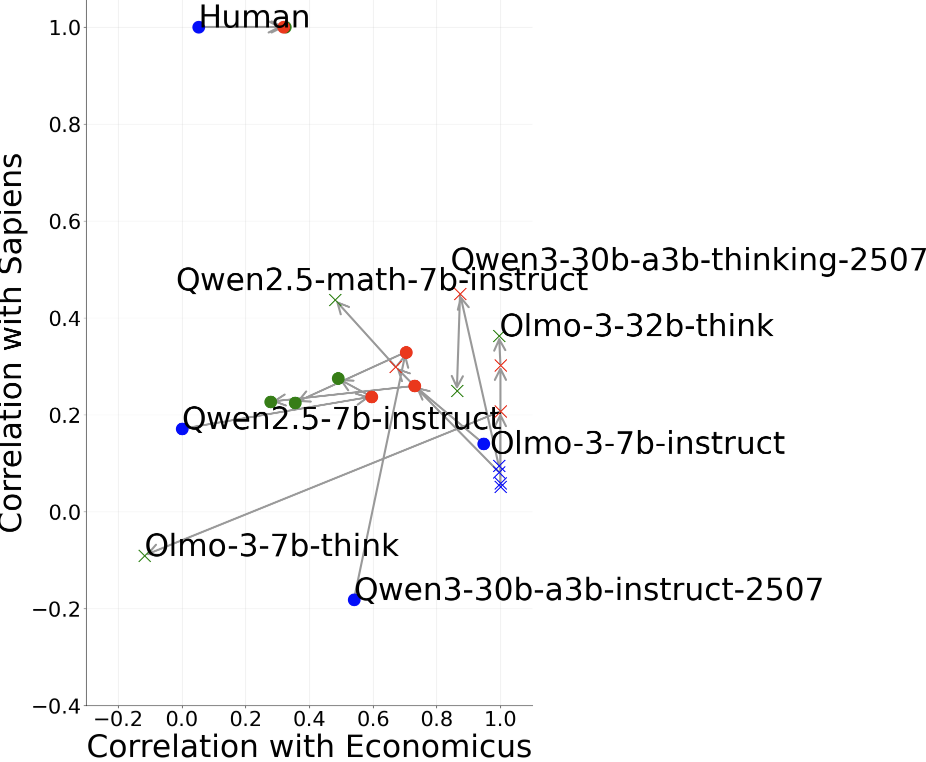}

    \caption{HE representation of the LLMs and the impact of sample size in implicit prospect. Blue: Explicit Prospect; Red: Sample size 100; Green: Sample size 20.}
    \label{fig:arrow_sample_size}
\end{figure}
In Figure~\ref{fig:arrow_sample_size}, we observe that the blue points (explicit prospects) are concentrated toward the bottom right, which is consistent with the picture when the implicit prospect results are aggregated. Only Qwen2.5-7B-instruct appears as an outlier. Furthermore, treating the explicit prospect as a limiting case with sample size 
$n=\infty$, frontier CMs exhibit a consistent pattern: as the sample size decreases, the models become less \ec-like and more human-like. In contrast, frontier RMs show smaller changes across explicit versus implicit prospects and between sample sizes ($n=20,100$).

\subsection{Additional Details on Impact of Training and Scale for Open Models}
\label{subsec:training_scale}

Here, we present the complete results for the sequential training checkpoints—SFT, DPO, and RLVF (final)—for both the instruction and thinking variants of the open \text{Olmo3} models. Their training details can be found in the technical report~\cite{olmo3_2025}.
The results for the dual-beta model, reported in Tables~\ref{tb:pm_training_stages} and~\ref{tb:pm_training_stages_experience}, do not indicate any systematic variation attributable to either the DPO or RLVF training stages. Again, given that \text{Olmo3-7B-Instruct-SFT} is initialized from \text{Olmo3-7B-Think-SFT}, the observed parameter differences are consistent with the hypothesis that the initial SFT stage may be the primary source of divergence between the two model variants.

\begin{table}[h!]
\centering
\begin{minipage}{0.5\textwidth}
\centering
\resizebox{\textwidth}{!}{%
\begin{tabular}{|l|c|c|c|c|c|c|}
\hline
\textbf{Model} 
& $\boldsymbol{\sigma}$ 
& $\boldsymbol{\gamma}$ 
& $\boldsymbol{\beta}_{\text{gain}}$ 
& $\boldsymbol{\beta}_{\text{loss}}$ 
& \textbf{Corr}
& \textbf{MSE} \\
\hline

Olmo-3-7B-Instruct-SFT & 0.88 & 0.80 & 309 & 130 & 0.80 & 0.029 \\
Olmo-3-7B-Instruct-DPO & 0.88 & 0.81 & 320 & 142 & 0.82 & 0.030 \\
Olmo-3-7B-Instruct & 0.88 & 0.81 & 215 & 121 & 0.73 & 0.041 \\\hline
Olmo-3-7B-Think-SFT & 1.27 & 1.13 & 1000 & 509 & 1.00 & 0.001 \\
Olmo-3-7B-Think-DPO & 1.33 & 1.16 & 1000 & 1000 & 1.00 & 0.000 \\
Olmo-3-7B-Think & 1.33 & 1.16 & 1000 & 1000 & 1.00 & 0.001 \\\hline
Olmo-3-32B-Think-SFT & 1.33 & 1.16 & 1000 & 1000 & 1.00 & 0.000 \\
Olmo-3-32B-Think-DPO & 1.33 & 1.16 & 1000 & 1000 & 1.00 & 0.000 \\
Olmo-3-32B-Think & 1.33 & 1.16 & 1000 & 1000 & 1.00 & 0.000 \\
\hline
\end{tabular}
}
\end{minipage}
\caption{Dual-beta prospect theory model parameters for models over sequential training stages (explicit prospects).}
\label{tb:pm_training_stages}
\end{table}

\begin{table}[h!]
\centering
\begin{minipage}{0.5\textwidth}
\centering
\resizebox{\textwidth}{!}{%
\begin{tabular}{|l|c|c|c|c|c|c|}
\hline
\textbf{Model} 
& $\boldsymbol{\sigma}$ 
& $\boldsymbol{\gamma}$ 
& $\boldsymbol{\beta}_{\text{gain}}$ 
& $\boldsymbol{\beta}_{\text{loss}}$ 
& \textbf{Corr}
& \textbf{MSE} \\
\hline

Olmo-3-7B-Instruct-SFT & 2.40 & 1.26 & 1.41 & 0.01 & 0.35 & 0.021 \\
Olmo-3-7B-Instruct-DPO & 1.50 & 1.05 & 2.86 & 0.07 & 0.37 & 0.020 \\
Olmo-3-7B-Instruct & 1.25 & 0.99 & 4.80 & 1.29 & 0.44 & 0.020 \\\hline
Olmo-3-7B-Think-SFT & 0.78 & 1.04 & 21.7 & 10.6 & 0.61 & 0.097 \\
Olmo-3-7B-Think-DPO & 0.98 & 0.86 & 1000 & 824 & 0.96 & 0.015 \\
Olmo-3-7B-Think & 0.73 & 0.71 & 1000 & 466 & 1.00 & 0.001 \\\hline
Olmo-3-32B-Think-SFT & 0.88 & 0.95 & 418 & 355 & 0.95 & 0.023 \\
Olmo-3-32B-Think-DPO & 0.96 & 0.89 & 1000 & 563 & 0.99 & 0.003 \\
Olmo-3-32B-Think & 0.95 & 1.07 & 1000 & 235 & 0.93 & 0.032 \\
\hline
\end{tabular}
}
\end{minipage}
\caption{Dual-beta prospect theory model parameters for models over sequential training stages (implicit prospects).}
\label{tb:pm_training_stages_experience}
\end{table}

\subsection{Confidence Intervals}
\label{subsec:ci_results}

Tables~\ref{tab:ci_b_e},~\ref{tab:ci_b_i},~\ref{tab:ci_o_e}, and~\ref{tab:ci_o_i} present the 95\% confidence intervals for the dual-beta prospect theory model, estimated via the parametric bootstrap method described in Appendix~\ref{subsubsec:model_fitting} with $B=1000$ iterations.

\begin{table}[h!]
    \centering
    \begin{minipage}{0.5\textwidth}
    \centering
    \resizebox{\textwidth}{!}{%
    \begin{tabular}{|l|c|c|c|c|}
        \hline
        \textbf{Model} & \textbf{$\boldsymbol{\sigma}$} & \textbf{$\boldsymbol{\gamma}$} & \textbf{$\boldsymbol{\beta}_{\text{gain}}$} & \textbf{$\boldsymbol{\beta}_{\text{loss}}$} \\
        \hline
        DeepSeek-R1 & [0.81, 0.96] & [0.80, 0.91] & [832, 832] & [711, 711] \\
        Gemini-2.5-Pro & [0.76, 0.86] & [0.81, 0.88] & [951, 951] & [1000, 1000] \\
        GPT-5.1 & [1.28, 1.35] & [1.14, 1.22] & [1000, 1000] & [1000, 1000] \\\hline
        Claude-4.5-Haiku & [0.88, 0.91] & [0.79, 0.81] & [221, 221] & [244, 244] \\
        DeepSeek-Chat & [0.84, 0.89] & [0.81, 0.84] & [181, 181] & [149, 149] \\
        Gemini-2.5-Flash & [0.84, 3.27] & [0.81, 4.61] & [4.53, 303] & [15, 1000] \\
        GPT-4.1 & [0.52, 0.95] & [1.33, 2.48] & [202, 282] & [0.01, 42] \\\hline
        Human & [0.32, 3.59] & [0.01, 2.21] & [7, 379] & [0.01, 189] \\
        \hline
    \end{tabular}
    }
    \end{minipage}
    \caption{95\% Confidence Intervals for dual-beta prospect theory parameters of black-box models with explicit prospects.}
    \label{tab:ci_b_e}
\end{table}

\begin{table}[h!]
    \centering
    \begin{minipage}{0.5\textwidth}
    \centering
    \resizebox{\textwidth}{!}{%
    \begin{tabular}{|l|c|c|c|c|}
        \hline
        \textbf{Model} & \textbf{$\boldsymbol{\sigma}$} & \textbf{$\boldsymbol{\gamma}$} & \textbf{$\boldsymbol{\beta}_{\text{gain}}$} & \textbf{$\boldsymbol{\beta}_{\text{loss}}$} \\
        \hline
        DeepSeek-R1 & [0.97, 1.01] & [0.98, 1.04] & [1000, 1000] & [346, 346] \\
        Gemini-2.5-Pro & [0.96, 0.99] & [0.97, 1.02] & [147, 147] & [100, 100] \\
        GPT-5.1 & [0.99, 1.03] & [0.96, 1.02] & [460, 460] & [1000, 1000] \\\hline
        Claude-4.5-Haiku & [1.01, 1.14] & [0.92, 1.10] & [3.27, 6.59] & [19.1, 28.1] \\
        DeepSeek-Chat & [0.68, 0.96] & [0.68, 0.78] & [8.43, 14.0] & [0.01, 1.24] \\
        Gemini-2.5-Flash & [1.46, 1.68] & [0.77, 0.89] & [1.29, 3.89] & [11.8, 17.2] \\
        GPT-4.1 & [1.44, 1.59] & [0.72, 0.77] & [0.01, 1.57] & [34.6, 47.5] \\\hline
        Human & [0.78, 1.19] & [0.79, 1.77] & [3.68, 11.1] & [3.71, 11.7] \\
        \hline
    \end{tabular}
    }
    \end{minipage}
    \caption{95\% Confidence Intervals for dual-beta prospect theory parameters of black-box models with implicit prospects.}
    \label{tab:ci_b_i}
\end{table}

\begin{table}[h!]
    \centering
    \begin{minipage}{0.5\textwidth}
    \centering
    \resizebox{\textwidth}{!}{%
    \begin{tabular}{|l|c|c|c|c|}
        \hline
        \textbf{Model} & \textbf{$\boldsymbol{\sigma}$} & \textbf{$\boldsymbol{\gamma}$} & \textbf{$\boldsymbol{\beta}_{\text{gain}}$} & \textbf{$\boldsymbol{\beta}_{\text{loss}}$} \\
        \hline
        Qwen2.5-7B-Instruct & [0.30, 0.90] & [0.35, 0.78] & [4.86, 140] & [0.01, 16.2] \\
        Qwen2.5-Math-7B & [0.80, 0.96] & [0.81, 0.98] & [1000, 1000] & [403, 403] \\
        Qwen3-30B-Instruct & [0.88, 2.06] & [0.81, 3.40] & [15.9, 213] & [0.01, 17.2] \\
        Qwen3-30B-Thinking & [0.82, 0.92] & [0.80, 0.90] & [726, 726] & [1000, 1000] \\\hline
        Olmo-3-7B-Instruct-SFT & [0.86, 0.89] & [0.80, 0.81] & [309, 309] & [130, 130] \\
        Olmo-3-7B-Instruct-DPO & [0.87, 0.89] & [0.80, 0.82] & [320, 320] & [142, 142] \\
        Olmo-3-7B-Instruct & [0.86, 0.89] & [0.80, 0.82] & [215, 215] & [121, 121] \\\hline
        Olmo-3-7B-Think-SFT & [0.93, 1.71] & [0.85, 1.58] & [1000, 1000] & [509, 509] \\
        Olmo-3-7B-Think-DPO & [1.28, 1.35] & [1.14, 1.22] & [1000, 1000] & [1000, 1000] \\
        Olmo-3-7B-Think & [1.28, 1.35] & [1.13, 1.22] & [1000, 1000] & [1000, 1000] \\\hline
        Olmo-3-32B-Think-SFT & [1.28, 1.35] & [1.13, 1.22] & [1000, 1000] & [1000, 1000] \\
        Olmo-3-32B-Think-DPO & [1.27, 1.34] & [1.14, 1.23] & [1000, 1000] & [1000, 1000] \\
        Olmo-3-32B-Think & [1.28, 1.35] & [1.14, 1.22] & [1000, 1000] & [1000, 1000] \\\hline
    \end{tabular}
    }
    \end{minipage}
    \caption{95\% Confidence Intervals for dual-beta prospect theory parameters of open-weight models with explicit prospects.}
    \label{tab:ci_o_e}
\end{table}

\begin{table}[h!]
    \centering
    \begin{minipage}{0.5\textwidth}
    \centering
    \resizebox{\textwidth}{!}{%
    \begin{tabular}{|l|c|c|c|c|}
        \hline
        \textbf{Model} & \textbf{$\boldsymbol{\sigma}$} & \textbf{$\boldsymbol{\gamma}$} & \textbf{$\boldsymbol{\beta}_{\text{gain}}$} & \textbf{$\boldsymbol{\beta}_{\text{loss}}$} \\
        \hline
        Qwen2.5-7B-Instruct & [23.4, 198] & [0.47, 0.90] & [0.01, 0.29] & [2.07, 3.28] \\
        Qwen2.5-Math-7B & [0.20, 0.77] & [0.76, 1.48] & [4.77, 9.88] & [0.01, 2.02] \\
        Qwen3-30B-Instruct & [1.02, 1.18] & [0.78, 0.89] & [13.8, 20.2] & [1.04, 4.84] \\
        Qwen3-30B-Thinking & [0.99, 1.00] & [0.87, 0.88] & [668, 668] & [585, 585] \\\hline
        Olmo-3-7B-Instruct-SFT & [1.45, 54.5] & [0.63, 5.88] & [0.62, 3.47] & [0.01, 0.76] \\
        Olmo-3-7B-Instruct-DPO & [1.19, 4.52] & [0.75, 3.52] & [1.40, 4.91] & [0.01, 1.26] \\
        Olmo-3-7B-Instruct & [1.08, 1.66] & [0.80, 2.07] & [3.10, 7.07] & [0.01, 2.82] \\\hline
        Olmo-3-7B-Think-SFT & [0.68, 0.92] & [0.89, 1.61] & [16.5, 30.5] & [8.02, 16.0] \\
        Olmo-3-7B-Think-DPO & [0.97, 1.18] & [0.85, 0.99] & [1000, 1000] & [824, 824] \\
        Olmo-3-7B-Think & [0.47, 1.11] & [0.55, 0.97] & [1000, 1000] & [466, 466] \\\hline
        Olmo-3-32B-Think-SFT & [0.78, 0.92] & [0.92, 1.12] & [418, 418] & [355, 355] \\
        Olmo-3-32B-Think-DPO & [0.93, 1.42] & [0.88, 1.45] & [1000, 1000] & [563, 563] \\
        Olmo-3-32B-Think & [0.91, 1.00] & [1.00, 1.39] & [1000, 1000] & [235, 235] \\\hline
    \end{tabular}
    }
    \end{minipage}
    \caption{95\% Confidence Intervals for dual-beta prospect theory parameters of open-weight models with implicit prospects.}
    \label{tab:ci_o_i}
\end{table}

\subsection{Ablation and Alternative Model Results}
\label{subsec:baseline_results}

This section presents results for alternative behavioral models defined in Appendix~\ref{subsubsec:model_fitting}: two restricted variants of the standard PT parameterization ($\sigma, \lambda, \gamma, \beta$) and a regret aversion model.

\subsubsection{Parameter Estimates: Restricted Prospect Theory Models}
\label{subsubsec:restricted_pt}
Table~\ref{tab:restricted_params_b_e} and~\ref{tab:restricted_params_b_i} detail the fitted parameters for the prospect theory ablation models. For Model 1, we report only $\beta$ (fixing $\sigma=\lambda=\gamma=1$). For Model 2, we report the shape parameters (fixing $\beta=1$).

\begin{table}[h!]
    \centering
    \begin{minipage}{0.5\textwidth}
    \centering
    \resizebox{\textwidth}{!}{%
    \begin{tabular}{|l|c|ccc|}
        \hline
        & \textbf{Model 1} & \multicolumn{3}{c|}{\textbf{Model 2 (Shape-Only)}} \\
        \textbf{Model} & $\boldsymbol{\beta}$ & $\boldsymbol{\sigma}$ & $\boldsymbol{\lambda}$ & $\boldsymbol{\gamma}$ \\
        \hline
        DeepSeek-R1 & 1000 & 391 & 3.19 & 1.35 \\
        Gemini-Pro & 745 & 0.74 & 1000 & 0.86 \\
        GPT-5.1 & 1000 & 338 & 4.13 & 1.34 \\\hline
        Claude Haiku & 14 & 277 & 1.26 & 1.21 \\
        DeepSeek-Chat & 95 & 294 & 1.41 & 1.49 \\
        Gemini-Flash & 18 & 14 & 2.03 & 1.39 \\
        GPT-4.1 & 0.01 & 0.81 & 70 & 0.69 \\\hline
        Human & 0.01 & 0.63 & 4.42 & 0.36 \\
        \hline
    \end{tabular}
    }
    \end{minipage}
    \caption{Restricted prospect theory parameters of black-box models for decisions with explicit prospects.}
    \label{tab:restricted_params_b_e}
\end{table}

\begin{table}[h!]
    \centering
    \begin{minipage}{0.5\textwidth}
    \centering
    \resizebox{\textwidth}{!}{%
    \begin{tabular}{|l|c|ccc|}
        \hline
        & \textbf{Model 1} & \multicolumn{3}{c|}{\textbf{Model 2 (Shape-Only)}} \\
        \textbf{Model} & $\boldsymbol{\beta}$ & $\boldsymbol{\sigma}$ & $\boldsymbol{\lambda}$ & $\boldsymbol{\gamma}$ \\
        \hline
        DeepSeek-R1 & 407 & 1.24 & 506 & 1.32 \\
        Gemini-Pro & 127 & 1.29 & 132 & 1.39 \\
        GPT-5.1 & 1000 & 1.32 & 1000 & 1.35 \\\hline
        Claude Haiku & 11 & 0.38 & 13 & 12 \\
        DeepSeek-Chat & 6 & 0.21 & 10 & 25 \\
        Gemini-Flash & 9 & 1.56 & 13 & 0.81 \\
        GPT-4.1 & 14 & 0.02 & 1000 & 0.15 \\\hline
        Human & 6 & 0.47 & 4.57 & 2.30 \\
        \hline
    \end{tabular}
    }
    \end{minipage}
    \caption{Restricted prospect theory parameters of black-box models for decisions with implicit prospects.}
    \label{tab:restricted_params_b_i}
\end{table}

\subsubsection{Parameter Estimates: Regret Aversion}
\label{subsubsec:ra}
Table~\ref{tab:regret_params_b_e} and~\ref{tab:regret_params_b_i} present the fitted parameters for Model 4, the regret aversion model. Here, $\lambda_{reg}$ represents decisiveness in the regret framework, while $\kappa$ and $\alpha$ control the shape of the regret function.

We observe that the regret aversion model does consistently poorly compared to the prospect theory model used in the main analysis. In terms of goodness-of-fit, the regret model yields significantly lower correlations and higher MSE across both prospect representations.

\begin{table}[h!]
    \centering
    \begin{minipage}{0.5\textwidth}
    \centering
    \resizebox{\textwidth}{!}{%
    \begin{tabular}{|l|c|c|c|c|c|}
        \hline
        \textbf{Model} & $\boldsymbol{\lambda_{reg}}$ & $\boldsymbol{\kappa}$ & $\boldsymbol{\alpha}$ 
        & \textbf{Corr} & \textbf{MSE} \\
        \hline
        DeepSeek-R1 & 0.01 & 108 & 0 & 0.53 & 0.14 \\
        Gemini-Pro & 0.01 & 87 & 0 & 0.45 & 0.15 \\
        GPT-5.1 & 0.01 & 144 & 0 & 0.61 & 0.15 \\\hline
        Claude Haiku & 0.01 & 82 & 0 & 0.58 & 0.07 \\
        DeepSeek-Chat & 0.01 & 61 & 0 & 0.48 & 0.08 \\
        Gemini-Flash & 0.01 & 127 & 0.31 & 0.77 & 0.03 \\
        GPT-4.1 & 0.01 & 994 & 538 & -0.26 & 0.10 \\\hline
        Human & 0.01 & 981 & 81 & -0.24 & 0.01 \\
        \hline
    \end{tabular}
    }
    \end{minipage}
    \caption{Regret aversion parameters of black-box models for decisions with explicit prospects.}
    \label{tab:regret_params_b_e}
\end{table}

\begin{table}[h!]
    \centering
    \begin{minipage}{0.5\textwidth}
    \centering
    \resizebox{\textwidth}{!}{%
    \begin{tabular}{|l|c|c|c|c|c|}
        \hline
        \textbf{Model} & $\boldsymbol{\lambda_{reg}}$ & $\boldsymbol{\kappa}$ & $\boldsymbol{\alpha}$ 
        & \textbf{Corr} & \textbf{MSE} \\
        \hline
        DeepSeek-R1 & 0.01 & 51 & 0 & 0.30 & 0.19 \\
        Gemini-Pro & 0.01 & 46 & 0 & 0.30 & 0.17 \\
        GPT-5.1 & 0.01 & 59 & 0 & 0.33 & 0.21 \\\hline
        Claude Haiku & 0.02 & 780 & 181 & 0.00 & 0.11 \\
        DeepSeek-Chat & 0.01 & 0 & 2.25 & -0.11 & 0.07 \\
        Gemini-Flash & 0.81 & 0 & 1.32 & 0.25 & 0.07 \\
        GPT-4.1 & 0.01 & 0 & 2.41 & -0.03 & 0.17 \\\hline
        Human & 0.06 & 825 & 24 & 0.02 & 0.03 \\
        \hline
    \end{tabular}
    }
    \end{minipage}
    \caption{Regret aversion parameters of black-box models for decisions with implicit prospects.}
    \label{tab:regret_params_b_i}
\end{table}

\subsection{Standard Prospect Theory Results}
\label{subsec:old_pt_results}

As discussed in Appendix~\ref{subsubsec:pt_model}, we also fit a standard prospect theory parameterization with four parameters: $\sigma$ (risk preference), $\lambda$ (loss aversion), $\gamma$ (probability weighting), and $\beta$ (decisiveness). While this formulation is widely used, $\lambda$ and $\beta$ are not jointly identifiable in pure-loss prospects, as the model can only recover their product (see Section~\ref{subsubsec:pt_model} for details). We report these results here for completeness; qualitative conclusions are consistent with the dual-beta specification reported in the main text.

\subsubsection{Frontier Models}

\begin{table}[h!]
\centering
\begin{minipage}{0.5\textwidth}
\centering
\resizebox{\textwidth}{!}{%
\begin{tabular}{|l|c|c|c|c|c|c|}
\hline
\textbf{Model} 
& $\boldsymbol{\sigma}$ 
& $\boldsymbol{\lambda}$ 
& $\boldsymbol{\gamma}$ 
& $\boldsymbol{\beta}$ 
& \textbf{Corr}
& \textbf{MSE} \\
\hline
DeepSeek-R1   & 0.79 & 1000 & 0.86 & 1000 & 0.98 & 0.012 \\
Gemini-Pro    & 0.82 & 0.57 & 0.83 & 1000 & 0.95 & 0.019 \\
GPT-5.1       & 1.33 & 1000 & 1.16 & 1000 & 1.00 & 0.002 \\
\hline
Claude Haiku  & 0.92 & 3.89 & 0.83 & 89  & 0.72 & 0.063 \\
DeepSeek-Chat & 0.84 & 0.43 & 0.84 & 159 & 0.75 & 0.042 \\
Gemini-Flash  & 0.95 & 1.58 & 0.94 & 20  & 0.87 & 0.015 \\
GPT-4.1       & 0.51 & 0.01 & 1.30 & 150 & 0.53 & 0.053 \\
\hline
Human         & 0.04 & 66   & 0.13 & 0.49 & 0.67 & 0.004 \\
\hline
\end{tabular}
}
\end{minipage}
\caption{Standard prospect theory parameters for frontier models with explicit prospects.}
\label{tab:old_pt_b_e}
\end{table}

\begin{table}[h!]
\centering
\begin{minipage}{0.5\textwidth}
\centering
\resizebox{\textwidth}{!}{%
\begin{tabular}{|l|c|c|c|c|c|c|}
\hline
\textbf{Model} 
& $\boldsymbol{\sigma}$ 
& $\boldsymbol{\lambda}$ 
& $\boldsymbol{\gamma}$ 
& $\boldsymbol{\beta}$ 
& \textbf{Corr}
& \textbf{MSE} \\
\hline
DeepSeek-R1   & 1.05 & 1.20 & 0.92 & 1000 & 0.99 & 0.005 \\
Gemini-Pro    & 0.84 & 1.05 & 1.24 & 128  & 0.95 & 0.021 \\
GPT-5.1       & 1.04 & 1000 & 0.94 & 1000 & 1.00 & 0.001 \\
\hline
Claude Haiku  & 0.70 & 0.90 & 2.10 & 12   & 0.66 & 0.059 \\
DeepSeek-Chat & 0.80 & 0.48 & 2.51 & 9    & 0.59 & 0.039 \\
Gemini-Flash  & 1.31 & 0.91 & 0.67 & 17   & 0.69 & 0.037 \\
GPT-4.1       & 1.12 & 3.42 & 0.64 & 18   & 0.69 & 0.080 \\
\hline
Human         & 1.03 & 0.89 & 0.83 & 9    & 0.55 & 0.022 \\
\hline
\end{tabular}
}
\end{minipage}
\caption{Standard prospect theory parameters for frontier models with implicit prospects.}
\label{tab:old_pt_b_i}
\end{table}

\subsubsection{Open Models}

\begin{table}[h!]
\centering
\begin{minipage}{0.5\textwidth}
\centering
\resizebox{\textwidth}{!}{%
\begin{tabular}{|l|c|c|c|c|c|c|}
\hline
\textbf{Model} 
& $\boldsymbol{\sigma}$ 
& $\boldsymbol{\lambda}$ 
& $\boldsymbol{\gamma}$ 
& $\boldsymbol{\beta}$ 
& \textbf{Corr}
& \textbf{MSE} \\
\hline
Qwen2.5-7B-Instruct   & 0.58 & 0.80 & 3.59 & 8    & 0.83 & 0.017 \\
Qwen2.5-Math-7B       & 1.11 & 0.34 & 1.03 & 1000 & 0.99 & 0.004 \\
Qwen3-30B-Instruct    & 0.94 & 0.01 & 0.86 & 48   & 0.57 & 0.041 \\
Qwen3-30B-Thinking    & 0.87 & 1000 & 0.81 & 1000 & 0.99 & 0.003 \\
\hline
Olmo-3-7B-Instruct    & 0.87 & 0.20 & 0.82 & 220  & 0.73 & 0.042 \\
Olmo-3-7B-Think       & 1.33 & 1000 & 1.15 & 1000 & 1.00 & 0.001 \\
Olmo-3-32B-Think      & 1.33 & 1000 & 1.16 & 1000 & 1.00 & 0.000 \\
\hline
\end{tabular}
}
\end{minipage}
\caption{Standard prospect theory parameters for open models with explicit prospects.}
\label{tab:old_pt_o_e}
\end{table}

\begin{table}[h!]
\centering
\begin{minipage}{0.5\textwidth}
\centering
\resizebox{\textwidth}{!}{%
\begin{tabular}{|l|c|c|c|c|c|c|}
\hline
\textbf{Model} 
& $\boldsymbol{\sigma}$ 
& $\boldsymbol{\lambda}$ 
& $\boldsymbol{\gamma}$ 
& $\boldsymbol{\beta}$ 
& \textbf{Corr}
& \textbf{MSE} \\
\hline
Qwen2.5-7B-Instruct   & 1.60 & 0.21 & 3.17 & 19  & 0.58 & 0.056 \\
Qwen2.5-Math-7B       & 0.55 & 0.12 & 0.88 & 7   & 0.54 & 0.027 \\
Qwen3-30B-Instruct    & 1.11 & 0.19 & 0.81 & 17  & 0.50 & 0.047 \\
Qwen3-30B-Thinking    & 0.98 & 1.67 & 0.85 & 913 & 0.95 & 0.024 \\
\hline
Olmo-3-7B-Instruct    & 1.25 & 0.27 & 1.00 & 4.80 & 0.44 & 0.020 \\
Olmo-3-7B-Think       & 0.82 & 0.69 & 0.78 & 702  & 1.00 & 0.001 \\
Olmo-3-32B-Think      & 0.76 & 1000 & 1.31 & 405  & 0.95 & 0.023 \\
\hline
\end{tabular}
}
\end{minipage}
\caption{Standard prospect theory parameters for open models with implicit prospects.}
\label{tab:old_pt_o_i}
\end{table}

\section{Human Instructions} \label{sec:human_instructions}
As all the participants were recruited via Prolific, they were provided informed consent prior to participation. The consent form described the nature of the study, the types of data collected, and how the data would be used and stored. Participants were informed that their responses would be recorded in anonymized form. Once they consented, they were presented with the following instructions prior to completing the task. Then in every page, they were presented with a question at a time with two options A,B. 

\begin{quote}
We invite you to participate in a research study being conducted by investigators from ***.

If you have any questions about the research study itself, please contact: ***.  
If you have questions, concerns, or complaints about your rights as a research participant, please contact: ***.

Thank you very much for your consideration of this research study.
\end{quote}

\begin{quote}
In each of the following 6 questions, you will be provided with two options with different payoffs and uncertainties. Please choose the one you prefer.
\end{quote}

\end{document}